\documentclass[lettersize,journal]{IEEEtran}
\usepackage{amsmath,amsfonts}
\usepackage{array}
\usepackage[caption=false,font=normalsize,labelfont=sf,textfont=sf]{subfig}
\usepackage{textcomp}
\usepackage{stfloats}
\usepackage{url}
\usepackage{verbatim}
\usepackage{graphicx}
\usepackage{cite}
\usepackage{booktabs}
\newcommand{\tabincell}[2]{\begin{tabular}{@{}#1@{}}#2\end{tabular}}
\usepackage{color}
\usepackage{amssymb}
\usepackage[ruled,vlined,linesnumbered]{algorithm2e}
\usepackage{multirow}

\begin{document}

\title{Towards Universal Modal Tracking with Online Dense Temporal Token Learning}

\def\tracker{ODTrack}
\def\modaltracker{UM-ODTrack}

\author{Yaozong~Zheng,~
        Bineng~Zhong$^{*}$,~
        Qihua~Liang,~
        Shengping~Zhang,~
        Guorong~Li,~
        
        Xianxian~Li,~
        Rongrong~Ji,~\IEEEmembership{Senior~Member,~IEEE}~

\thanks{Yaozong Zheng, Bineng Zhong, Qihua Liang, and Xianxian Li are with the Key Laboratory of Education Blockchain and Intelligent Technology, Ministry of Education, and the Guangxi Key Laboratory of Multi-Source Information Mining and Security, Guangxi Normal University, Guilin 541004, China. (E-mail: yaozongzheng@stu.gxnu.edu.cn, bnzhong@gxnu.edu.cn, qhliang@gxnu.edu.cn, lixx@gxnu.edu.cn)}
\thanks{Shengping Zhang is currently a Professor with the School of Computer Science and Technology, Harbin Institute of Technology, Weihai, Shandong 264209, China. (E-mail: s.zhang@hit.edu.cn)}
\thanks{Guorong Li is currently an Associate Professor with the University of Chinese Academy of Sciences. (E-mail: liguorong@ucas.ac.cn)}
\thanks{Rongrong Ji is currently a Professor with Media Analytics and Computing Lab, Department of Artificial Intelligence, School of Informatics, Xiamen University, 361005, China. (E-mail: rrji@xmu.edu.cn)}
\thanks{(Corresponding author: Bineng Zhong.)}
}

\markboth{Journal of \LaTeX\ Class Files,~Vol.~14, No.~8, August~2021}%
{Shell \MakeLowercase{\textit{et al.}}: A Sample Article Using IEEEtran.cls for IEEE Journals}


\maketitle

\begin{abstract}
We propose a universal video-level modality-awareness tracking model with online dense temporal token learning (called {\modaltracker}). It is designed to support various tracking tasks, including RGB, RGB+Thermal, RGB+Depth, and RGB+Event, utilizing the same model architecture and parameters. Specifically, our model is designed with three core goals: \textbf{Video-level Sampling}. We expand the model's inputs to a video sequence level, aiming to see a richer video context from an near-global perspective. \textbf{Video-level Association}. Furthermore, we introduce two simple yet effective online dense temporal token association mechanisms to propagate the appearance and motion trajectory information of target via a video stream manner. \textbf{Modality Scalable}. We propose two novel gated perceivers that adaptively learn cross-modal representations via a gated attention mechanism, and subsequently compress them into the same set of model parameters via a one-shot training manner for multi-task inference. This new solution brings the following benefits: (i) The purified token sequences can serve as temporal prompts for the inference in the next video frames, whereby previous information is leveraged to guide future inference. (ii) Unlike multi-modal trackers that require independent training, our one-shot training scheme not only alleviates the training burden, but also improves model representation. Extensive experiments on visible and multi-modal benchmarks show that our {\modaltracker} achieves a new \textit{SOTA} performance. The code will be available at \url{https://github.com/GXNU-ZhongLab/ODTrack}.
\end{abstract}


\section{Introduction} \label{sec:introduction}

\IEEEPARstart{V}{isual} tracking aims to uniquely identify and track an object within a video sequence by using arbitrary target queries.
It is an important and challenging task in computer vision. Over the past few years, with significant advancements in the visual tracking technologies, numerous real-world applications have emerged, including video understanding \cite{videochat}, autonomous driving \cite{autodriving} and human-computer interaction \cite{Hand}, etc.

\begin{figure*}
      \centering
      \includegraphics[width=1\linewidth]{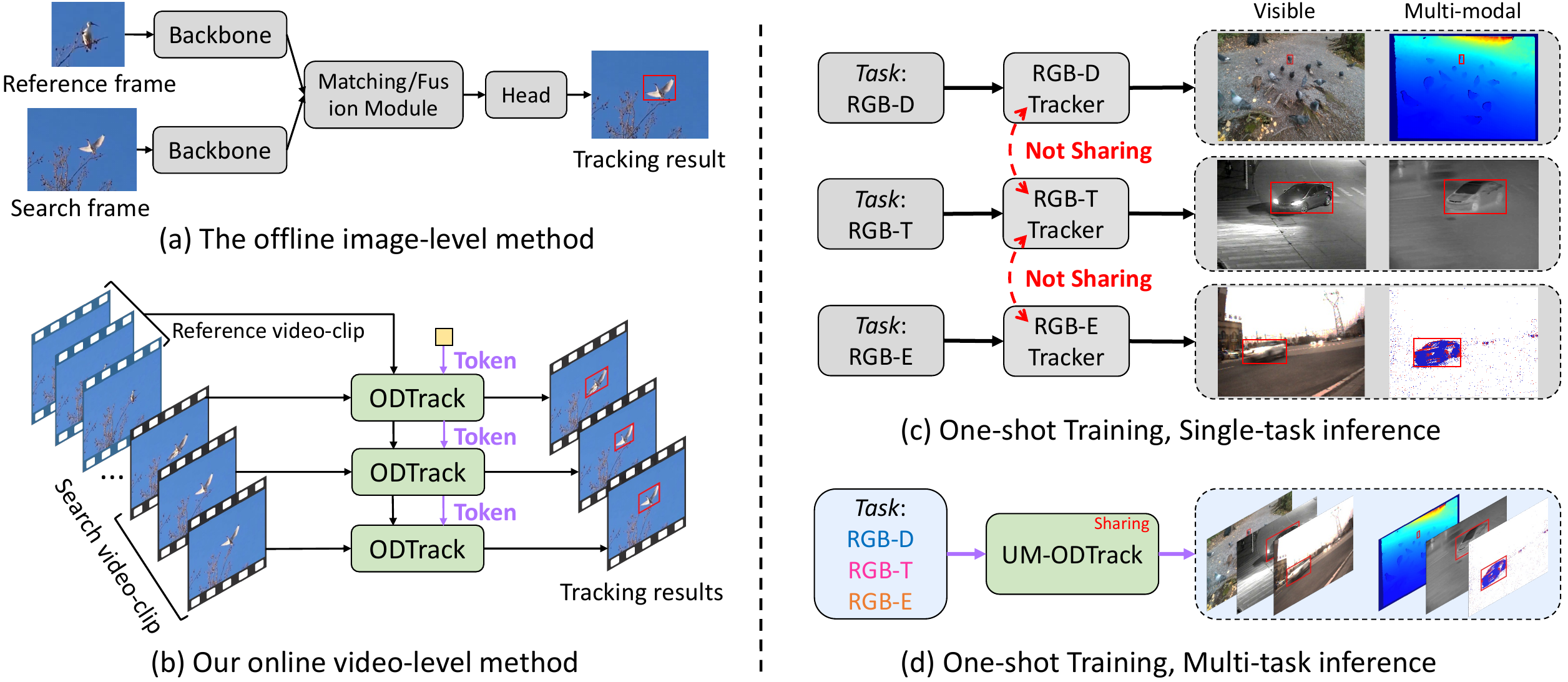}
       \caption{Comparison of tracking methods.
       (a) The offline image-level tracking methods \cite{SiamRPN++,transt} based on sparse sampling and image-pair matching.
       (b) Our online video-level tracking method based on video sequence sampling and temporal token propagation.
       (c) The multi-modal tracking methods \cite{vipt,TBSI,CEUTrack,EMT} based on one-shot training and single-task inference (i.e., one-model to one-task, one-to-one).
       (d) Our universal modal-awareness tracking model based on one-shot training and multi-task inference (i.e., one-model to many-tasks, one-to-many).}
       \label{fig:motivation}
    \end{figure*}

Multi-modal visual perception has been a persistent research challenge in the tracking community. According to various data modalities, visual tracking is primarily categorized into sub-learning tasks, including RGB tracking \cite{SiamFC,SiamRPN++,Siamban,stark,ostrack,ARTrack}, RGB+Thermal tracking (RGB-T) \cite{mfdimp,TBSI,MPLT}, RGB+Depth tracking (RGB-D) \cite{DMTracker,RGBD1K,EMT}, and RGB+Event tracking (RGB-E) \cite{EFED,visevent,CEUTrack}.
In a pure RGB tracking task, its algorithms require modeling feature relationships among visible video frames to construct robust single-modal trackers. In multi-modal tracking tasks, their algorithms are primarily built upon RGB tracking models and require cross-modal understanding to achieve inter-modal complementarity. As shown in Fig.\ref{fig:motivation}, although the above two types of trackers perform well in their scenarios, their tracking performance is still unsatisfied due to the following modeling schemes, i.e., image-pair sampling, image-pair matching, and one-model to one-task learning.
    
(\textit{i}) \textbf{\textit{Image-pair sampling}}. As shown in Fig.\ref{fig:motivation}(a), previous efforts \cite{SiamRPN,SiamRN,transt,ostrack} have primarily relied on a sparse sampling strategy, such as using only one reference frame and one search frame, and required the trackers to interact independently within each image-pair. However, since visual tracking inherently contains rich temporal data, this simple sampling strategy falls short in accurately representing the motion state of an object, posing a significant challenge for trackers to analyze dynamic video content.

(\textit{ii}) \textbf{\textit{Image-pair matching}}. Traditional feature matching/fusion methods\cite{Siamban,Ocean,SiamGAT,SBT} focus on the target appearance similarity and thus overlook continuous cross-frame associations. Recently, some works \cite{stark,VideoTrack,mixformer,seqtrack} have attempted to enrich spatio-temporal information by adding multiple video frames (i.e., dynamic frames) and updating the model parameters \cite{DiMP50}, so as to simulate video modeling. However, they still fail to account for the cross search frame association of target information at the video level, limiting their spatio-temporal relationships to the range of selected video frames.
Meanwhile, to the best of our knowledge, video-level association is still rarely explored in the field of multi-modal tracking.

(\textit{iii}) \textbf{\textit{One-model to one-task learning}}. As shown in Fig.\ref{fig:motivation}(c), due to disparities in representations across different modalities, previous multi-modal tracking methods \cite{vipt,TBSI,CEUTrack,EMT} persistently lean on a one-model to one-task learning paradigm (i.e., \textbf{\textit{one-to-one}}). As a result, their compatibility is limited, i.e., their challenge lies in efficiently and economically adapting to various multi-modal tracking tasks. Thus, this \textit{one-to-one} tracking paradigm is constrained by customized networks with domain-specific knowledge, making it impractical to share the same model architectures and parameters for multi-task inference.

Based on the above observations, this inspires us to think: \textit{Could we design a universal video-level model to solve tracking tasks across various modalities in a unified manner?}
To achieve this goal, we introduce a novel universal video-level modality-awareness tracking model with online dense temporal token learning (called {\modaltracker}). It aims to integrate various sub-tracking tasks (i.e., RGB tracking, RGB-T tracking, RGB-D tracking, and RGB-E tracking) through a unified set of model architecture and parameters.
Specifically, we construct our {\modaltracker} with generic and high-performance from the following three aspects.

(\textit{i}) \textbf{From image-pair to video-level sampling}.
We treat each video sequence as a continuous sentence, enabling us to employ language modeling for a comprehensive contextual understanding of the video content.
Based on this idea, we expand the model inputs from an image-pair to a video stream level. This improvement contributes to address the limitations of a traditional sparse sampling strategy and enables the exploration of rich temporal dependencies.

(\textit{ii}) \textbf{From image-level to video-level association}. 
As shown in Fig.\ref{fig:motivation}(b), unlike conventional tracking methods \cite{stark,ostrack,mixformer,ARTrack}, we reformulate object tracking as a token sequence propagation task that densely associates the contextual relationships of across video frames in an auto-regressive manner. Under this new modeling paradigm, we design two simple yet effective temporal token association attention mechanisms (i.e., \textit{concatenated} and \textit{separated} temporal token attention) that capture the spatio-temporal trajectory relationships of a target instance using an online token propagation manner, thus allowing the processing of video-level inputs of greater length.

(\textit{iii}) \textbf{From single-modal to multi-modal awareness}. 
Furthermore, as illustrated in Fig.\ref{fig:motivation}(d), to abandon the traditional \textit{one-to-one} learning paradigm with a large training burden, we introduce a new universal modality-aware tracking pipeline. It utilizes a single model to simultaneously learn multiple sub-tracking tasks, aiming to improve the general modality representation capability (\textbf{\textit{one-to-many}}).
Specifically, we design two interesting gated perceivers (i.e., \textit{conditional gate} and \textit{gated modal-scalable perceiver}) that adaptively learn cross-modal representations through a gated attention mechanism.
Subsequently, our {\modaltracker} adopts a one-shot training scheme to optimize a unified multi-modal tracking model, aiding in compressing the learned latent representations into the same set of model parameters. According to our one-shot training scheme, {\modaltracker} can naturally learn a shared visual-semantic feature space while respecting the heterogeneity of different tasks, thereby achieving multi-task inference.

In summary, {\modaltracker} is a \textit{general-purpose}, \textit{task-flexible}, \textit{high-performance} tracking framework. Extensive experiments on seven visible (including LaSOT \cite{LaSOT}, TrackingNet \cite{TrackingNet}, GOT10K \cite{got10k}, LaSOT$_{\rm{ext}}$ \cite{lasot-ext}, VOT2020 \cite{VOT2020}, TNL2K \cite{tnl2k}, and OTB100 \cite{OTB2015}) and five multi-modal tracking benchmarks (including LasHeR \cite{lasher}, RGBT234 \cite{rgbt234}, DepthTrack \cite{depthtrack}, VOT-RGBD2022 \cite{VOT2022}, and VisEvent \cite{visevent}) show the effectiveness of our {\modaltracker}. The main contributions of our work are as follows:

\begin{itemize}
    \item To the best of our knowledge, we offer the first universal video-level modal-awareness tracking model for visual tracking community. {\modaltracker} needs to be trained only once and can realize multi-task inference, including RGB-T/D/E tracking tasks, using the same architecture and parameters.

    \item For video-level association, we introduce two temporal token propagation attention mechanisms that compress the discriminative features of the target into a token sequence. This token sequence serves as a prompt to guide the inference of future frames, thus avoiding complex online update strategies.

    \item For multi-modal awareness, we present two novel gated perceivers that adaptively learns latent representations across modalities, contributing to multi-task unified training and inference of our model.

\end{itemize}

This paper builds upon our conference paper \cite{ODTrack} and significantly extends it in various aspects.
First, we provide more in-depth discussions regarding motivations and implementations of our universal video-level multi-modal tracking algorithms.
Second, we build {\modaltracker} based on two novel gated perceivers to achieve universal multi-modal tracking with one model structure and parameters.
Third, more experiments and ablation studies are designed to demonstrate the effectiveness of the core components of our model, especially focusing on the techniques used in our video-level multi-modal tracking model.
Furthermore, more visual results are added for providing a comprehensive and intuitive understanding of the performance of our algorithms in multi-modal tracking scenarios.

\section{Related Work}

With the advancement of deep learning techniques \cite{lu2023tf,lu2024mace,lu2024robust,gao2024eraseanything,li2025set,he2024diffusion,he2025segment,he2023hqg,he2025unfoldir,he2025run,he2025reti,he2024weakly,xiao2024survey,he2023strategic,he2023camouflaged,he2023degradation,gong2021eliminate,gong2024cross,gong2022person,peng2025directing,peng2024lightweight,peng2025boosting,peng2025towards}, visual object tracking has attracted increasing attention from the research community.
According to various data modalities, visual tracking is primarily categorized into sub-learning tasks, including RGB tracking and multi-modal tracking (i.e., RGB-T/D/E tracking).

\subsection{RGB-based single-modal Tracking}

In the past, the popular trackers\cite{SiamFC,SiamRPN,DaSiamRPN,SiamMask,SiamRPN++,SiamDW,SiamFC++,SiamCAR,Siamban,Ocean,siamattn,SiamRN,transt,stark,SiamPIN} are dominated by the Siamese tracking paradigm, which achieves tracking by image-pair matching. 
To improve the accuracy and robustness of trackers, several different approaches are proposed. For example, 
SiamRPN \cite{SiamRPN} introduces region proposal network (RPN) \cite{FasterRCNN} into the Siamese framework, which solves the challenge of scale variation. 
SiamFC++ \cite{SiamFC++}, SiamBAN \cite{Siamban}, SiamCAR \cite{SiamCAR}, and Ocean \cite{Ocean} propose simple yet effective anchor-free trackers, which avoids redundant anchor settings. 
To address the inefficiency of cross-correlation operation, SiamGAT \cite{SiamGAT}, ACM \cite{ACM}, PG-Net \cite{PG-Net}, and TransT \cite{transt} design various improvement modules using the new convolutional operation or attention mechanism. 

In recent years, the introduction of the transformer \cite{attention} enables trackers \cite{stark,SBT,mixformer,simtrack,ostrack,mmtrack,SSTrack} to explore more powerful and deeper feature interactions, resulting in significant advances in tracking algorithm development. For example, 
inspired by the idea of DETR \cite{DETR}, STARK \cite{stark} designs an end-to-end tracking framework based on the encoder-decoder architecture.
Furthermore, OSTrack \cite{ostrack}, SimTrack \cite{simtrack}, and Mixformer \cite{mixformer} propose new tracking models based on transformer backbone networks \cite{vit,CvT}, which greatly improve the tracking performance.

However, most of these methods are designed based on sparse image-pair sampling strategy.
With this design paradigm, the tracker struggles to accurately comprehend the object’s motion state in the temporal dimension and can only resort to traditional Siamese similarity for appearance modeling.
In contrast to these approaches, we reformulate object tracking as a token sequence propagation task and aim to extend traditional tracker to efficiently exploit target temporal information in an auto-regressive manner.

\subsection{Multi-Modal Tracking} 

Although RGB trackers \cite{ostrack,mixformer} perform well in most common tracking scenarios, some challenging scenarios, such as occlusion and low visibility, reveal the problem of unreliable tracking.
To alleviate this problem, multi-modal tracking with auxiliary modality (i.e., thermal, depth, or event modalities) aggregation becomes a promising solution.

In the field of RGB+Thermal tracking, the thermal infrared cues contributes to improving tracking capabilities in low-light or dark environments by measuring the thermal radiation distribution of a target object. 
mfDiMP \cite{mfdimp} leverages other algorithm to generate thermal data and introduces a multi-level fusion mechanism, including pixel-level, feature-level, and response-level fusion, to construct an end-to-end RGB-T tracking model.
APFNet \cite{APFNet} designs five attribute-specific fusion branches that integrate RGB and thermal features corresponding to different challenges for robust RGB-T tracking.
DMCNet \cite{DMCNet} proposes a duality-gated mutual condition network to enhance target representations in RGB and TIR modalities, especially in the presence of low-quality modalities.
TBSI \cite{TBSI} exploits templates to bridge the cross-modal interaction between RGB and TIR search regions by gathering and distributing target-relevant objects and environment contexts.

In the field of RGB+Depth tracking, the RGB-D trackers, leveraging depth maps, can discern the characteristics of a target instance, contributing to achieving robust tracking in scenarios involving occlusion. 
Yan et al. \cite{depthtrack} propose a new depth tracking dataset (i.e., DepthTrack) and an RGB-D tracker (i.e., DeT) based on ATOM \cite{ATOM} or DiMP \cite{DiMP50}, to unlock the potential of depth tracking.
Recently, to address the dataset deficiency issue, Zhu et al. \cite{RGBD1K} propose another new large-scale RGBD tracking dataset (i.e., RGBD1K) and create a new RGBD tracker (i.e., SPT) based on STARK \cite{stark}.

In the field of RGB+Event tracking, the event cameras enable the RGB-E trackers to provide more accurate target localization in fast-moving scenarios by generating the event stream. 
To simplify the model structure and computational complexity, Tang et al. \cite{CEUTrack} propose a single-stage RGBE tracker (i.e., CEUTrack) and a new RGBE dataset (i.e., COESOT) for color-event unified tracking.
ProTrack \cite{protrack} and ViPT \cite{vipt} employ a prompt fine-tuning learning strategy to set specific visual cues for various multi-modal tracking, adapting the frozen pre-trained RGB tracker to multi-modal tracking tasks.
Furthermore, Un-Track \cite{UnTrack} makes the first attempt to learn representations of multiple modalities using low-rank factorization and reconstruction techniques and unify them into the same set of model parameters.

The above-mentioned studies, while proficient in performing multi-modal tracking tasks, are limited in their cross-task inference capabilities due to their one-model to one-task modeling style (i.e., \textbf{one-to-one}).
In contrast to these efforts, we propose a universal modality-awareness tracking model (called {\modaltracker}) based on two interesting gated perceivers that enables the compression of latent representations of depth, thermal infrared, and event modalities into a unified model architecture and parameters (i.e., one-model to many-tasks, \textbf{one-to-many}). Therefore, our method significantly improves the representational capabilities and reduces expensive training resources.

\subsection{Online Temporal Modeling Tracking}

To explore temporal cues within the Siamese framework, several online update methods are carefully designed.
UpdateNet\cite{UpdateNet} introduces an adaptive updating strategy, which utilizes a custom network to fuse accumulated templates and generate a weighted updated template feature for visual tracking. DCF-based trackers\cite{ATOM,DiMP50,PrDiMP-50} excel at updating model parameters online using sophisticated optimization techniques, thereby improving the robustness of the tracker. STMTrack\cite{STMTrack} and TrDiMP\cite{trdimp} employ attention mechanism to effectively extract contextual information along the temporal dimension. STARK\cite{stark} and Mixformer\cite{mixformer} specifically design target quality branch for updating template frame, which aids in improving the tracking results.
Recently, there has been a gradual surge in research attention towards modeling temporal context from various perspectives.
TCTrack \cite{TCTrack} introduces an online temporal adaptive convolution and an adaptive temporal transformer that aggregates temporal contexts at two levels containing feature extraction and similarity map refinement. VideoTrack \cite{VideoTrack} designs a new tracker based on video transformer and uses a simple feedforward network to encode temporal dependencies.
ARTrack \cite{ARTrack} presents a new time-autoregressive tracker that estimates the coordinate sequence of an object progressively.
    
Nevertheless, the above tracking algorithms still suffer from the following limitations: (1) The optimization process is complex, involving the design of specialized loss functions\cite{DiMP50}, multi-stage training strategies\cite{stark}, and manual update rules\cite{stark}, and (2) Although they explore temporal information to some extent, they fail to investigate \textit{how temporal cues propagate across search frames}.

In this work, we introduce a new dense context association mechanism from a token propagation perspective, which offers a solution to circumvent intricate optimization processes and training strategies. Furthermore, we propose a new baseline approach, which focuses on unlocking the potential of temporal modeling through the propagation of target motion trajectory information.

\section{Approach}

\subsection{Architecture Design}
In this section, we introduce our universal video-level modal-awareness framework called {\modaltracker}, which supports various tracking tasks, including RGB, RGB+Thermal, RGB+Depth, and RGB+Event tracking.

\begin{figure*}
      \centering
      \includegraphics[width=0.95\linewidth]{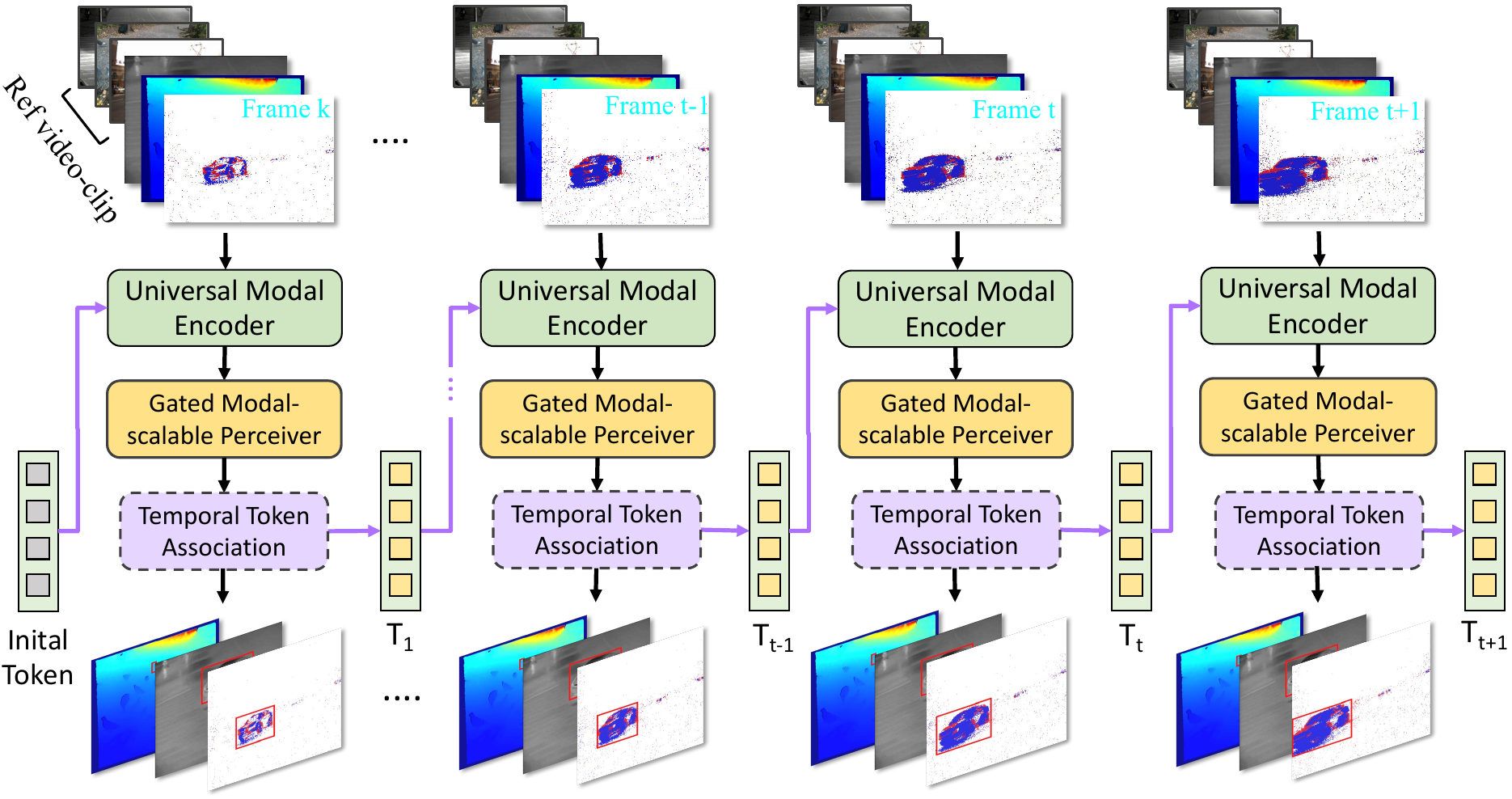}
       \caption{\textbf{{\modaltracker} Framework Architecture}.
       The {\modaltracker} takes video clips, consisting of reference and search frames, of various modalities and given length as input. Then, the model utilizes a temporal token association attention mechanism to generate a temporal token for each video frame. These temporal tokens are subsequently propagated to the following frames in an auto-regressive manner, enabling cross-frame propagation of target trajectory information.
       }
       \label{fig:framework}
    \end{figure*}

\begin{figure*}
      \centering
      \includegraphics[width=1\linewidth]{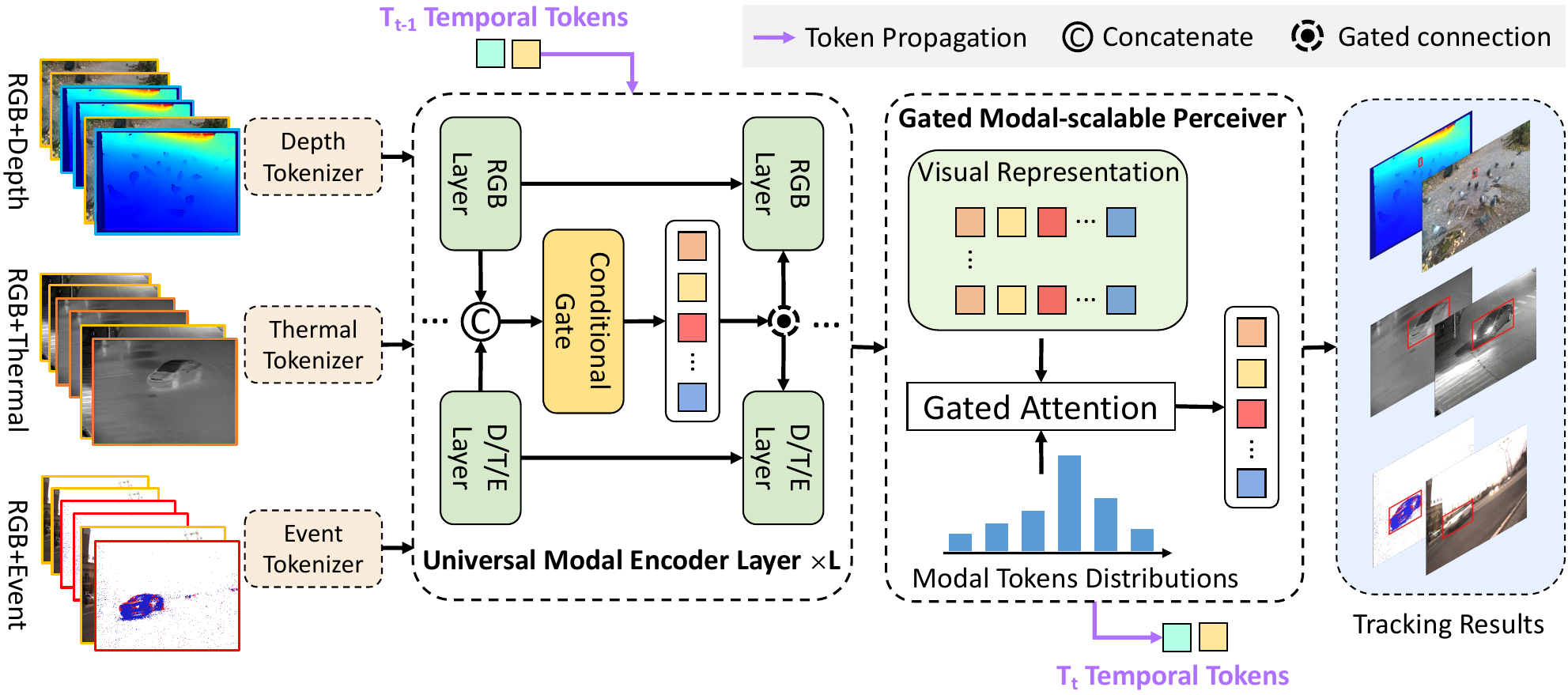}
       \caption{\textbf{{\modaltracker} for Multi-modal Tracking}.
       Our {\modaltracker} comprises three modality tokenizers, a shared universal modal encoder containing RGB encoder and D/T/E encoder, two gated perceivers including a conditional gate and a gated modal-scalable perceiver, and a prediction head. The tokenizers and the universal modality encoder jointly process inputs from three different modalities and collaboratively learn unified multi-modal representations. Subsequently, the two gated perceivers adaptively aggregate cross-modal knowledge by evaluating the relative importance of each modality. The gated connection represents selective multi-modal fusion based on a gating mechanism.}
       \label{fig:modal_framework}
    \end{figure*}

Fig.\ref{fig:modal_framework} and Fig.\ref{fig:framework} give an overview of our {\modaltracker} framework for video-level multi-modal tracking.
Specifically, we model an entire video as a continuous sequence, and decode the localization of a target instance frame by frame in an auto-regressive manner.
Firstly, we present a novel video sequence sampling strategy designed specifically to meet the input requirements of the video-level model (Principle 1: \textit{video-level sampling}).
Then, we propose a novel modality tokenizer that tokenizes different modality sources in a shared encoding manner.
Subsequently, to capture the spatio-temporal trajectory information of the target instance within the video sequences, we introduce two simple yet effective temporal token association attention mechanisms (Principle 2: \textit{video-level association}).
Furthermore, we introduce two powerful gated perceivers to adaptively learn universal visual representations across modalities, thereby improving model generalization across different tracking scenarios (Principle 3: \textit{modality-scalable}).

Based on the above modeling techniques, we will obtain a universal modal-awareness tracking model that can inference multiple sub-tracking tasks simultaneously using the same model architecture and parameters.
Detailed descriptions are presented in the following sections.

\subsection{Video-level Multi-modal Tracking Formulation}

Our focus lies in constructing a universal video-level multi-modal tracking framework. To provide a comprehensive understanding of our {\modaltracker} framework, it is pertinent to describe the concept of video-level multi-modal tracking.

First, we provide a review of previously prominent image-pair matching tracking methodologies \cite{SiamFC,SiamRPN,SiamRPN++,Siamban,transt}. Given a pair of video frames, i.e., a reference frame $R \in \mathbb{R}^{3 \times H_r \times W_r}$ and a search frame $S \in \mathbb{R}^{3 \times H_s \times W_s}$, the mainstream visual trackers $\Psi$ are formulated as 
   \begin{equation}
     B = \Psi(R, S),
     \label{eq:define}
   \end{equation}
where $B$ denotes the predicted box coordinates of the current search frame. If $\Psi$ is a conventional convolutional siamese tracker\cite{SiamRPN++,Siamban,transt}, it undergoes three stages, namely feature extraction, feature fusion, and bounding box prediction. Whereas if $\Psi$ is a transformer tracker\cite{ostrack,mixformer,simtrack}, it consists solely of a backbone and a prediction head network, where the backbone integrates the processes of feature extraction and fusion.

Specifically, a transformer tracker receives a series of non-overlapping image patches (the resolution of each image patch is $p \times p$) as input. This means that a 2D reference-search image pair needs to pass through a patch embedding layer to generate multiple 1D image token sequences $\{f_r \in \mathbb{R}^{D \times N_r}, f_s \in \mathbb{R}^{D \times N_s} \}$, where $D$ is the token dimension, $N_r = H_rW_r / p^2$, and $N_s = H_sW_s / p^2$. These 1D image tokens are then concatenated and loaded into a $M$-layer transformer encoder for feature extraction and relationship modeling. 
Each transformer layer $\delta$ contains a multi-head attention and a multi-layer perceptron.
Here, we formulate the forward process of the $\ell^{th}$ transformer layer as follows:

   \begin{equation}
     f_{rs}^{\ell} = \delta^{\ell}(f_{rs}^{\ell-1}), \quad \quad \quad \ell=1,2,...,M
     \label{eq:backbone}
   \end{equation}
where $f_{rs}^{\ell-1}$ denotes the concatenated token sequence of the reference-search image pair generated from the $(\ell-1)^{th}$ transformer layer, and $f_{rs}^{\ell}$ represents the token sequence generated by the current $\ell^{th}$ transformer layer.

Using the above modeling approach, we can construct a concise and elegant tracker to achieve per-frame tracking. Nevertheless, this modeling approach has two clear drawbacks:
1) The created tracker solely focuses on intra-frame target matching and lacks the ability to establish inter-frame associations necessary for tracking object across a video stream.
2) The created tracker is limited to single-modal tracking scenarios and lacks the ability to rapidly extend to multi-modal tracking due to domain-specific knowledge bias. 
Consequently, these limitations hinder the research of video-level multi-modal tracking algorithms.

In this work, we aim to alleviate these challenges and propose a new design paradigm for universal video-level modal-awareness tracking algorithm.
First, we extend the inputs of the tracking framework from the image-pair level to the video level for temporal modeling. Then, we introduce a temporal token sequence $T$ designed to propagate information about the appearance, spatio-temporal location and trajectory of the target instance in a video sequence.
Formally, we formulate video-level tracking as follows:
   \begin{equation}
     B = \Psi(\{R\}_k, \{S\}_n, T),
     \label{eq:tracker}
   \end{equation}
where $\{R\}_k$ denotes the RGB reference frames of length $k$, and $\{S\}_n$ represents the RGB search frames of length $n$. With such a setting, a video-level tracking framework is constructed that receives video clip of arbitrary length to model spatio-temporal trajectory relationships of the target object.

Furthermore, to improve the universal modal-awareness capability of the video-level tracker, we extend it to the field of multi-modal tracking.
First, we extend the inputs from single-modal to multi-modal range. 
Next, a shared universal mdoal encoder containing RGB encoder and D/T/E encoder is used to extract and fuse the RGB video-clips and auxiliary video-clips, respectively.
Sequentially, two novel gated perceivers are designed to learn universal latent representations across modalities.
The definition is as follows:
   \begin{equation}
     B = \Psi(\{R, R^{'}\}_k, \{S, S^{'}\}_n, T),
     \label{eq:modaltracker}
   \end{equation}
where $\{R^{'}\}_k$ denotes the reference frames of length $k$ from auxiliary modal, and $\{S^{'}\}_n$ represents the search frames of length $n$ from auxiliary modal. $T^{'}$ is the temporal token from auxiliary modal. 
We describe the proposed core modules in more detail in the next section.

\subsection{Video Sequence Sampling Strategy}
Most existing trackers \cite{stark,mixformer,ostrack} commonly sample single-modal image-pairs within a short-term interval, such as 50, 100, or 200 frame intervals. However, this sampling approach poses a potential limitation as these trackers fail to capture the long-term motion variations of the tracked object, thereby constraining the robustness of tracking algorithms in long-term scenarios. Meanwhile, they cannot perceive the real-time status of the target from multiple modal perspectives.

To obtain richer multi-modal spatio-temporal trajectory information of the target instance from long-term video sequences, we deviate from the traditional short-term image-pair sampling method and propose a new video sequence sampling strategy.
Specifically, in the training phase, we establish a larger sampling interval and randomly extract multiple video frames within this interval to form video clips $( \{R, R^{'}\}_k, \{S, S^{'}\}_n )$ of any modality and any length.
Although this sampling approach may seem simplistic, it enables us to approximate the content of the entire video sequence. This is crucial for video-level multi-modal tracking modeling.

\subsection{Modality Tokenizers}
Intuitively, considering the variability in input frames from different modalities (i.e., depth, thermal infrared, and event), a conventional approach is to design separate tokenizers for each modality \cite{onellm}. This allows diverse input frames to be transformed into token vectors with the same sequential format.
On the contrary, considering the possibility of shared semantic information among different modalities, we regard depth, thermal infrared, and event data as a unified visual representation.
To be specific, a shared modality tokenizer is designed to uniformly transform data from different modalities into the same one-dimensional sequence.
For visual inputs containing diverse modality information such as depth, thermal infrared, and event, we employ a single 2D convolutional layer as the unified tokenizer. Subsequently, a transformer-based universal modal encoder is utilized to process these tokens.

\subsection{Gated Perceivers}

Due to the limited modality-awareness abilities of the fundamental visual tracker, once it is trained on RGB tracking benchmarks, it cannot readily adapt to complex multi-modal tracking scenarios.
Therefore, we design two simple yet effective modules, namely \textit{conditional gate} and \textit{gated modal-scalable perceiver}, as illustrated in Fig.\ref{fig:modal_framework}, to learn universal cross-modal representations adaptively.

\textbf{\textit{Conditional Gate}}. 
Aiming to implement multi-modal representation learning in a shared universal modal encoder, we add conditional gate modules between each encoder layer in a residual manner.
In a conditional gate module, the visible features and corresponding auxiliary features (i.e., depth, thermal, and event) are cross-modality aligned along the channel dimension to complement the abundant details from other modalities.
The aligned multi-modal representations are then gated by the conditional gate module to facilitate cross learning between modalities.

Formally, the conditional gate module can be normalized to the following equation:
   \begin{equation}
     \widehat{f} = gate(\sigma([f_t, f_{t}^{'}])) + f, \quad f \in \{ f_t, f_{t}^{'} \}
     \label{eq:conditional_gate}
   \end{equation}

where $f_t$ and $f_t^{'}$ represent the visible and auxiliary modal features extracted from the $t^{th}$ video frame under a specific modality. $\sigma(\cdot)$ is an embedding layer for scaling dimensions.
$gate(\cdot)$ is a gated network. It dynamically controls the representation learning for multi-modal tracking based on the quality between modal sources, which is evaluated by a two-layer perceptron and a gated activation function. $\widehat{f}$ stands for the output features of the conditional gate module.
Notably, the learning parameters of the last conditional gated network layer are initialized to zero, allowing its output to match that of the fundamental visual tracker, thus contributing to improving training stability.

\textbf{\textit{Gated Modal-scalable Perceiver (GMP)}}.
After executing the universal modal encoder, we can obtain a visible features $f_t$, an auxiliary features $f_t^{'}$, a visible temporal token sequence $T_t$, and an auxiliary modal temporal token sequence $T_t^{'}$.
Two temporal tokens from different modalities, with their feature space distributions reflecting the appearance and motion trajectory information of the same target object across multiple modal sources.
Therefore, we design a novel modal-scalable perceiver based on a gated attention mechanism to further enhance the perception of multi-modal tracking scenarios.
Specifically, the learned multi-modal representations are cross-attended with two temporal modality tokens to construct generic modality dependencies from multiple views. This multi-modal relationship can be indicated as the formula:
   \begin{equation}
     f_{p} = \textnormal{UM-Attn}(\sigma([f_t, f_t^{'}]), [T_t, T_t^{'}]),
     \label{eq:gate_attn}
   \end{equation}
   \begin{equation}
     {\widehat{f}_{p}} = gate(f_{p}) + f_{p}, 
     \label{eq:gate2}
   \end{equation}
   \begin{equation}
     {\widetilde{f}_{p}} = \textnormal{UM-MLP}(\widehat{f}_{p}), 
     \label{eq:gate_mlp}
   \end{equation}
where $\textnormal{UM-Attn}(\cdot,\cdot)$ denotes a multi-modal cross-attention layer with the former input as query and the later as key and value. $\textnormal{UM-MLP}(\cdot)$ represents a multi-modal feed forward network layer.
$f_{p}$ is the output features of $\textnormal{UM-Attn}$ operation in the GMP module.
$\widehat{f}_{p}$ is the output features of gate operation in the GMP module.
$\widetilde{f}_p$ stands for the output features of the GMP module. 
By employing this novel gated attention mechanism, our {\modaltracker} can adaptively aggregate multi-modal information into a shared visual-semantic feature space, effectively improving the modal awareness capability of our tracker, thereby achieving truly universal modal tracking for the first time.

\subsection{Temporal Token Association Attention Mechanism}

Instead of employing a complex video transformer \cite{VideoTrack} as the foundational framework for encoding video content, our design takes a new perspective by utilizing a simple 2D transformer architecture, i.e., 2D ViT \cite{vit}.

\begin{figure}[t]
      \centering
      \includegraphics[width=1\linewidth]{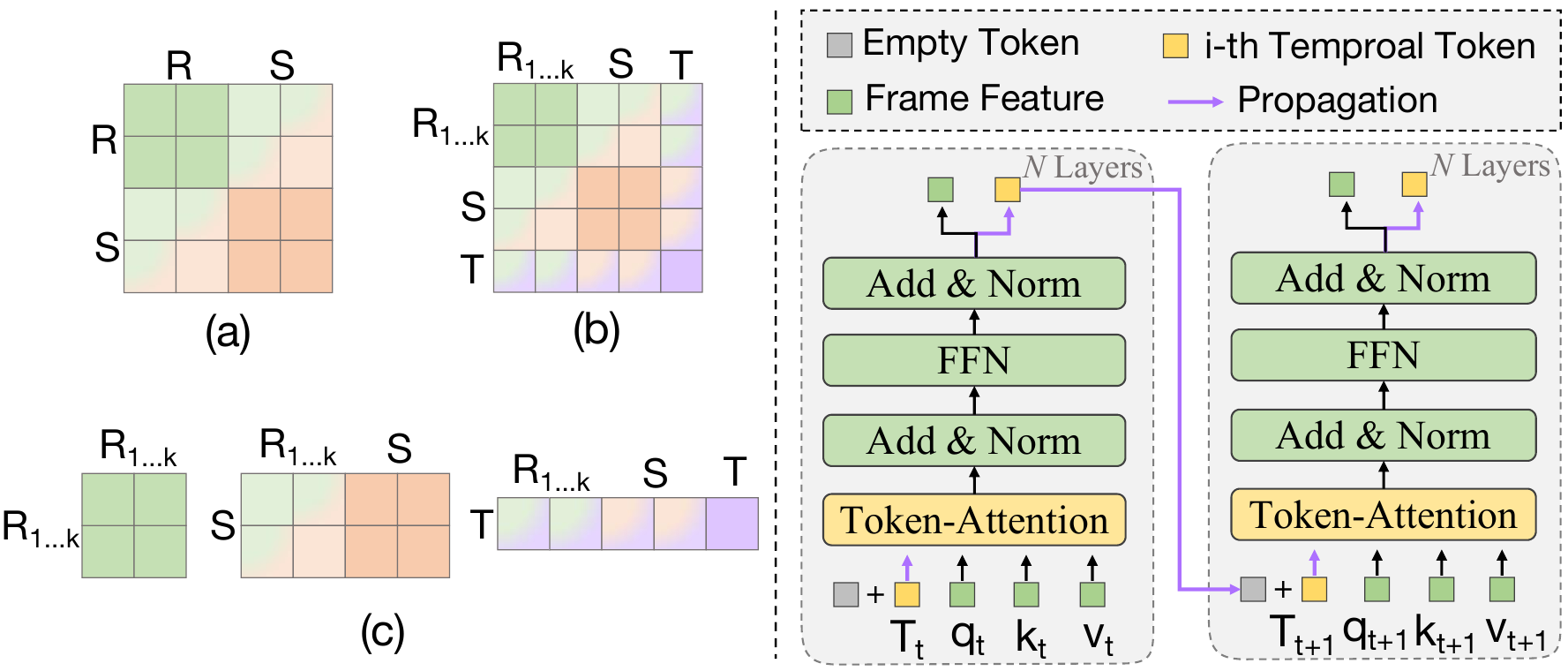}
       \caption{Left: the architecture of temporal token association attention mechanism. Right: illustration of online token propagation. (a) Original reference-search attention mechanism, (b) and (c) Different variants of the proposed temporal token association attention mechanisms. $R$ is a single reference frame, $R_{1...k}$ denotes  the reference frames of length $k$, $S$ represents the current search frame, and $T$ is the temporal token sequence of current video frame.
       }
       \label{fig:module}
    \end{figure}

To construct an elegant instance-level inter-frame correlation mechanism, it is imperative to extend the original 2D attention operations to extract and integrate video-level features.
In our approach, we design two temporal token attention mechanisms based on the concept of \textbf{\textit{compression-propagation}}, namely \textit{concatenated token attention mechanism} and \textit{separated token attention mechanism}, as shown in Fig.\ref{fig:module}(left).
The core design involves injecting additional information into the attention operations, such as more video sequence content and temporal token vectors, enabling them to extract richer spatio-temporal trajectory information of the target instance.

In Fig.\ref{fig:module}(a), the original attention operation commonly employs an image pair as inputs, where the process of modeling their relationships can be represented as $f=\textnormal{Attn}([R, S])$. In this paradigm, the tracker can only engage in independent interactions within each image pair, establishing limited temporal correlations. 
In Fig.\ref{fig:module}(b), the proposed concatenated token attention mechanism extends the input to the aforementioned video sequence, enabling dense modeling of spatio-temporal relationships across frames.
Inspired by the contextual nature of language formed through concatenation, we apply the concatenation operation to establish context for video sequences as well.
Its formula can be represented as:
   \begin{equation}
     \begin{split}
        f_t &= \textnormal{Attn}([\{R\}_k, S_t, T_t]) \\
        &= \sum_{s''t''} V_{s''t''} \cdot \frac{\exp \langle q_{st}, k_{s''t''} \rangle}{\sum_{s't'} \exp \langle q_{st}, k_{s't'} \rangle},
      \end{split}
     \label{eq:msa}
   \end{equation}
where $T_t$ is the temporal token sequence of $t^{th}$ video frame. $[\cdots,\cdots]$ denotes concatenation among tokens. $q_{st}$, $k_{st}$ and $v_{st}$ are spatio-temporal linear projections of the concatenated feature tokens.
On the other hand, when performing multi-modal tracking task, the current temporal token association attention mechanism is also applicable. Specifically, similar to the visible temporal token, the multi-modal temporal token $T_t^{'}$ is a vector initialized with zero that is used to extract the appearance and spatio-temporal localization information of a target instance in multi-modal tracking scenarios. The formula represented as follows:
   \begin{equation}
     \begin{split}
        f_t^{'} &= \textnormal{Attn}([\{R^{'}\}_k, S_t^{'}, T_t^{'}]).
      \end{split}
     \label{eq:modal_msa}
   \end{equation}

It is worth noting that we introduce a temporal token for each video frame, with the aim of storing the target trajectory information of the sampled video sequence. In other words, we \textbf{\textit{compress}} the current spatio-temporal trajectory information of the target into a token vector, which is used to \textbf{\textit{propagate}} to the subsequent video frames.

\begin{algorithm}[t]
\caption{Tracking with ours models}
\label{algo:UM_ODTrack}
\SetAlgoLined
 \KwIn{Reference frame $R_1$ and its ground-truth $B_1$ for any modality, Search frame $S$ for any modality}
 \KwOut{Tracking results $\left \{ B_{t=2:n} \right \}$ for any modality}

 $memory = []$ \tcp*[r]{init reference memory}
 
 $T, T^{'} = None, None$ \tcp*[r]{init temporal token}

 \For{task in \{RGB, RGB-D, RGB-T, RGB-E\}}{
     \For{$t = 2$ to $n$}{
     
     \If{not memory is None}{
        select reference frames $R$ at equal intervals;
     }
     
     $T_t = T$;

     \If{task in \{RGB-D, RGB-T, RGB-E\}}{
        $T_t^{'} = T^{'}$;
        
        Extract and fuse features for $(\{R, R^{'}\}_k, \{S, S^{'}\}_n, T, T^{'} )$ using Eq.(\ref{eq:conditional_gate})-(\ref{eq:modal_propagate});


        $B_t, T_{t+1}, T_{t+1}^{'} = \Psi(\{R, R^{'}\}_k, \{S, S^{'}\}_n, T, T^{'} )$;

        $T^{'} = T_{t+1}^{'}$;
     }
     \ElseIf{task is RGB}{
        Extract and fuse features for $(\{R\}_k, S_t, T_{t} )$ using Eq.(\ref{eq:propagate});

        $B_t, T_{t+1} = \Psi(\{R\}_k, S_t, T_{t} )$;
     }

     Crop $S_t$ based on $B_t$ yields a new $R_t$;
    
     Save $R_t$ to $memory$;
     
     $T = T_{t+1}$;
     }
 }
\end{algorithm}

Once the target information is extracted by the temporal token, we propagate the token vector from $t^{th}$ frame to $(t+1)^{th}$ frame in an auto-regressive manner, as shown in Fig.\ref{fig:module}(right).
Firstly, the $t^{th}$ temporal token $T_{t}$ is added to the $(t+1)^{th}$ empty token $T_{empty}$, resulting in an update of the content token $T_{t+1}$ for $(t+1)^{th}$ frame, which is then propagated as input to the subsequent frames. Formally, the propagation process for visible and multi-modal tracking is:
   \begin{equation}
     \begin{split}
        T_{t+1} &= T_t + T_{empty}, \\
        f_{t+1} &= \textnormal{Attn}([\{R\}_k, S_{t+1}, T_{t+1}]),
      \end{split}
     \label{eq:propagate}
   \end{equation}
   
   \begin{equation}
     \begin{split}
        T_{t+1}^{'} &= T_t^{'} + T_{empty}^{'}, \\
        f_{t+1}^{'} &= \textnormal{Attn}([\{R^{'}\}_k, S_{t+1}^{'}, T_{t+1}^{'}]).
      \end{split}
     \label{eq:modal_propagate}
   \end{equation}
where $T_t^{'}$ is the temporal token sequence of $t^{th}$ auxiliary modal video frame. $T_{empty}^{'}$ is the empty token of $(t+1)^{th}$ auxiliary modal video frame.

In this new design paradigm, we can employ temporal tokens as prompts for inferring the next frame, leveraging past information to guide future inference. Moreover, our model implicitly propagates appearance, localization, and trajectory information of the target instance through online token propagation. This significantly improves tracking performance of the video-level framework.

On the other hand, as illustrated in Fig.\ref{fig:module}(c), the proposed separated token attention mechanism decomposes attention operation into three sub-processes: self-information aggregation between reference frames, cross-information aggregation between reference and search frames, and cross-information aggregation between temporal token and video sequences. This decomposition improves the computational efficiency of the model to a certain extent, while the token association aligns with the aforementioned procedures.

\subsubsection{Discussions with Online Update.}

Most previous tracking algorithms combine online updating methods to train a spatio-temporal tracking model, such as adding an extra score quality branch \cite{stark} or an IoU prediction branch \cite{ATOM}. They typically require complex optimization processes and update decision rules. In contrast to these methods, we avoid complex online update strategies by utilizing online iterative propagation of token sequences, enabling us to achieve more efficient model representation and computation.

   \begin{figure}[t]
      \centering
      \includegraphics[width=1\linewidth]{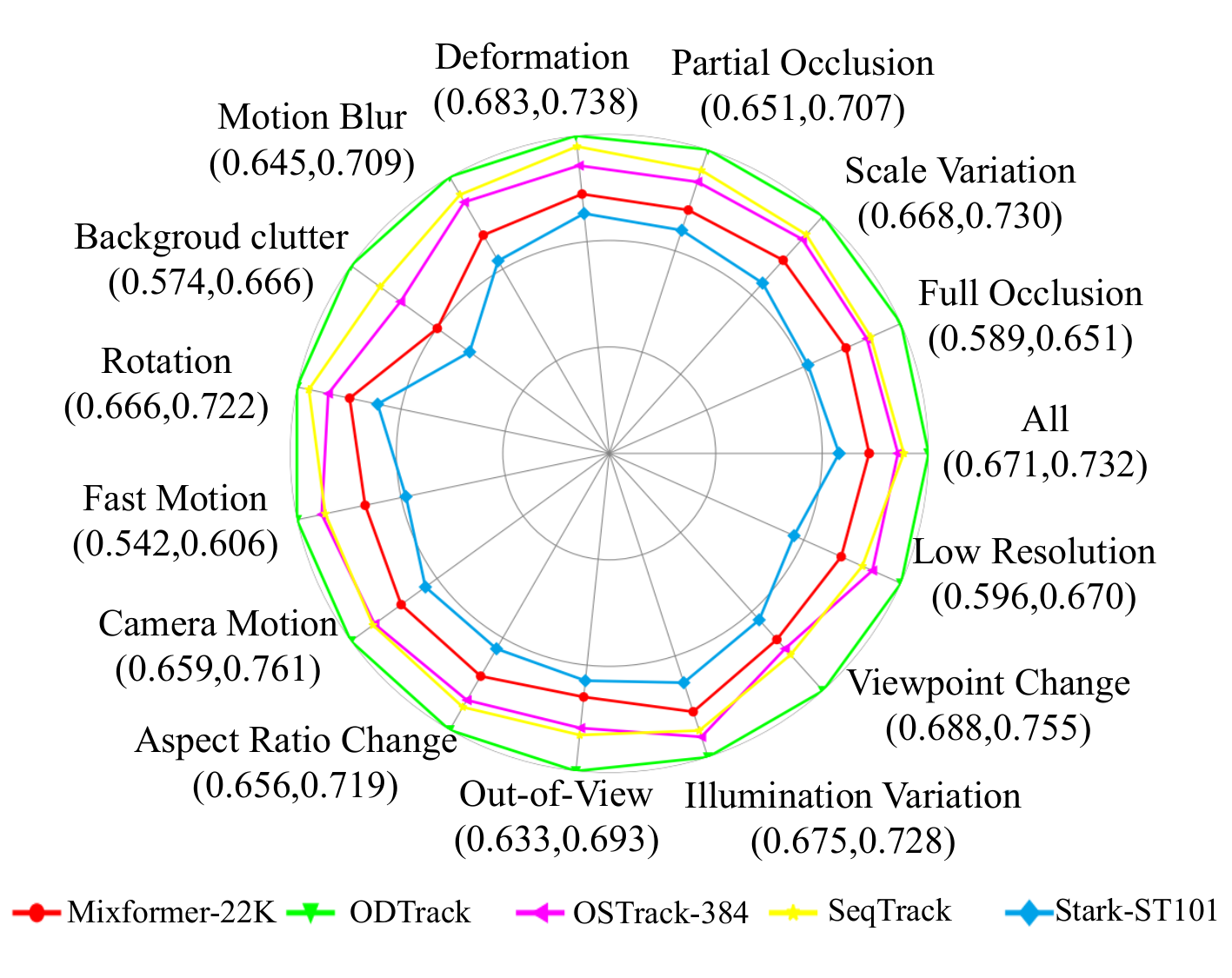}
       \caption{AUC scores of different attributes on LaSOT.}
       \label{fig:attrs}
    \end{figure}

\begin{figure}[t]
\begin{center}
\subfloat[Success Rate]{
\includegraphics[width=0.48\linewidth]{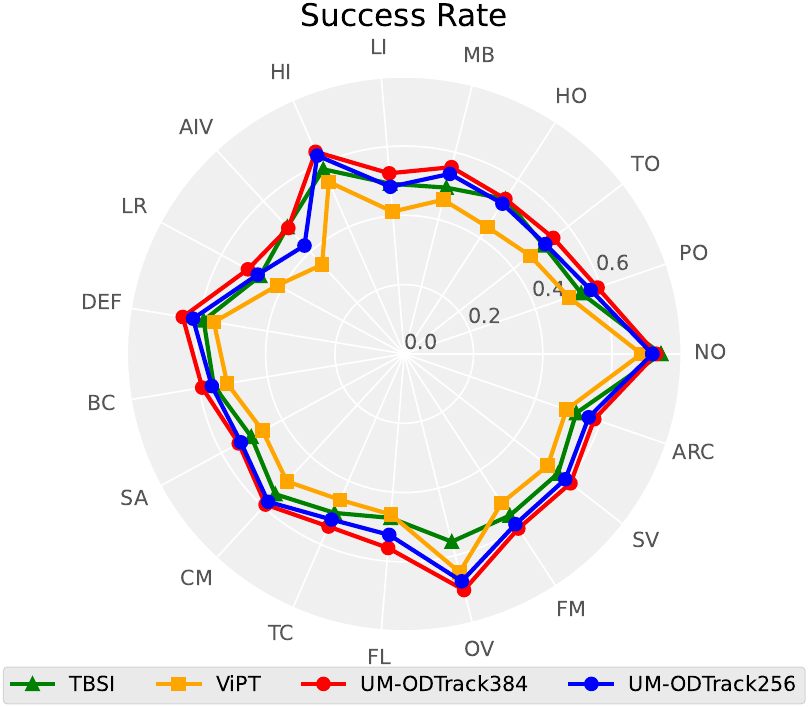}}
\subfloat[Precision Rate]{
\includegraphics[width=0.48\linewidth]{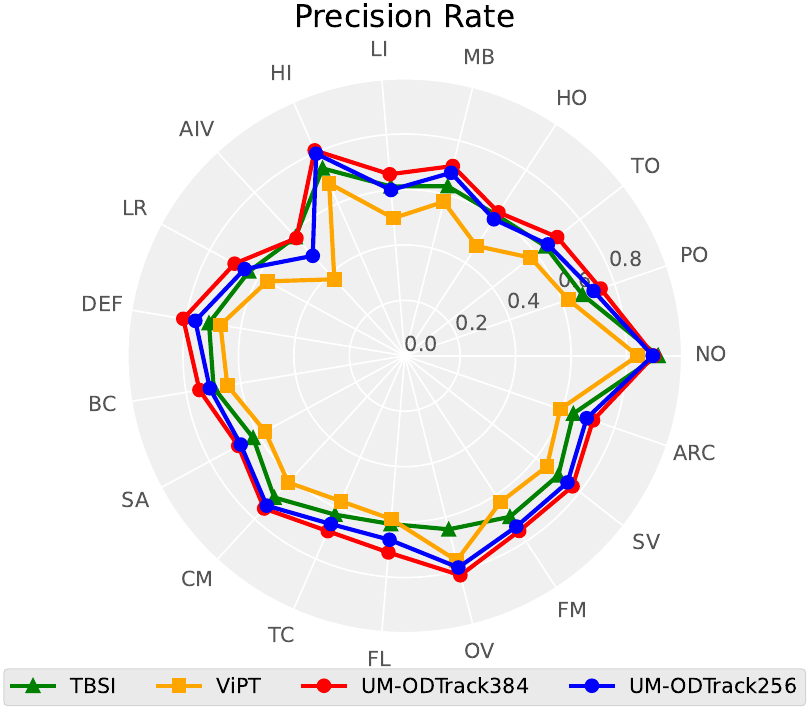}}
\end{center}
\caption{Success and precision rate of different attributes on LasHeR \cite{lasher}.}
\label{fig:lasher_attr}
\end{figure}

\subsection{One-shot Training and Universal Inference}

\textbf{Prediction Head.}
For the design of the prediction head network, similar to \cite{ostrack}, we employ conventional classification head and bounding box regression head to achieve the desired outcome.
The classification score map $\mathbb{R}^{1 \times \frac{H_s}{p} \times \frac{W_s}{p}}$, bounding box size $\mathbb{R}^{2 \times \frac{H_s}{p} \times \frac{W_s}{p}}$, and offset size $\mathbb{R}^{2 \times \frac{H_s}{p} \times \frac{W_s}{p}}$ for the prediction are obtained through three sub-convolutional networks, respectively.

\begin{table*}[t]
    \centering
    \caption{Comparison with state-of-the-arts on four popular benchmarks: GOT10K \cite{got10k}, LaSOT \cite{LaSOT}, TrackingNet \cite{TrackingNet}, and LaSOT$_{\rm{ext}}$ \cite{lasot-ext}. Where $*$ denotes for trackers only trained on GOT10K. The best two results are highlighted in {\color{red}red} and {\color{blue}blue}, respectively. The subscripts $256$ and $384$ represent configurations with search frames of $256 \times 256$ and $384 \times 384$ respectively.}
    \label{tab:results}
    \resizebox{\textwidth}{!}{
    \begin{tabular}{l|ccc|ccc|ccc|ccc}
    \toprule
     \multicolumn{1}{c|}{\multirow{2}{*}{Method}} & \multicolumn{3}{c|}{GOT10K$^*$} & \multicolumn{3}{c|}{LaSOT} & \multicolumn{3}{c|}{TrackingNet} & \multicolumn{3}{c}{LaSOT$_{\rm{ext}}$} \\
     \cline{2-13}
      & AO & SR${_{0.5}}$ & SR${_{0.75}}$ & AUC & P${_{\rm{Norm}}}$ & P & AUC & P${_{\rm{Norm}}}$ & P & AUC & P${_{\rm{Norm}}}$ & P \\
      \midrule
      SiamFC \cite{SiamFC} & 34.8 & 35.3 & 9.8 & 33.6 & 42.0 & 33.9 & 57.1 & 66.3 & 53.3 & 23.0 & 31.1 & 26.9 \\
      ECO\cite{ECO} & 31.6 & 30.9 & 11.1 & 32.4 & 33.8 & 30.1 & 55.4 & 61.8 & 49.2 & 22.0 & 25.2 & 24.0 \\
      ATOM \cite{ATOM} & 55.6 & 63.4 & 40.2 & 51.5 & 57.6 & 50.5 & 70.3 & 77.1 & 64.8 & 37.6 & 45.9 & 43.0 \\
      SiamRPN++ \cite{SiamRPN++} & 51.7 & 61.6 & 32.5 & 49.6 & 56.9 & 49.1 & 73.3 & 80.0 & 69.4 & 34.0 & 41.6 & 39.6 \\
      DiMP \cite{DiMP50} & 61.1 & 71.7 & 49.2 & 56.9 & 65.0 & 56.7 & 74.0 & 80.1 & 68.7 & 39.2 & 47.6 & 45.1 \\
      SiamRCNN \cite{siamrcnn} & 64.9 & 72.8 & 59.7 & 64.8 & 72.2 & - & 81.2 & 85.4 & 80.0 & - & - & - \\
      Ocean \cite{Ocean} & 61.1 & 72.1 & 47.3 & 56.0 & 65.1 & 56.6 & - & - & - & - & - & - \\
      STMTrack \cite{STMTrack} & 64.2 & 73.7 & 57.5 & 60.6 & 69.3 & 63.3 & 80.3 & 85.1 & 76.7 & - & - & - \\
      TrDiMP \cite{trdimp} & 67.1 & 77.7 & 58.3 & 63.9 & - & 61.4 & 78.4 & 83.3 & 73.1 & - & - & - \\
      TransT \cite{transt} & 67.1 & 76.8 & 60.9 & 64.9 & 73.8 & 69.0 & 81.4 & 86.7 & 80.3 & - & - & - \\
      Stark \cite{stark} & 68.8 & 78.1 & 64.1 & 67.1 & 77.0 & - & 82.0 & 86.9 & - & - & - & - \\
      KeepTrack \cite{keeptrack} & - & - & - & 67.1 & 77.2 & 70.2 & - & - & - & 48.2 & - & - \\
      SBT-B \cite{SBT} & 69.9 & 80.4 & 63.6 & 65.9 & - & 70.0 & - & - & - & - & - & - \\
      Mixformer \cite{mixformer} & 70.7 & 80.0 & 67.8 & 69.2 & 78.7 & 74.7 & 83.1 & 88.1 & 81.6 & - & - & - \\
      TransInMo \cite{TransInMo} & - & - & - & 65.7 & 76.0 & 70.7 & 81.7 & - & - & - & - & - \\
      OSTrack$_{256}$ \cite{ostrack} & 71.0 & 80.4 & 68.2 & 69.1 & 78.7 & 75.2 & 83.1 & 87.8 & 82.0 & 47.4 & 57.3 & 53.3 \\
      OSTrack$_{384}$ \cite{ostrack} & 73.7 & 83.2 & 70.8 & 71.1 & 81.1 & 77.6 & 83.9 & 88.5 & 83.2 & 50.5 & 61.3 & 57.6 \\
      AiATrack \cite{aiatrack} & 69.6 & 80.0 & 63.2 & 69.0 & 79.4 & 73.8 & 82.7 & 87.8 & 80.4 & 47.7 & 55.6 & 55.4 \\
      SeqTrack$_{256}$ \cite{seqtrack} & 74.7 & 84.7 & 71.8 & 69.9 & 79.7 & 76.3 & 83.3 & 88.3 & 82.2 & 49.5 & 60.8 & 56.3 \\
      SeqTrack$_{384}$ \cite{seqtrack} & 74.5 & 84.3 & 71.4 & 71.5 & 81.1 & 77.8 & 83.9 & 88.8 & 83.6 & 50.5 & 61.6 & 57.5 \\
      GRM \cite{GRM} & 73.4 & 82.9 & 70.4 & 69.9 & 79.3 & 75.8 & 84.0 & 88.7 & 83.3 & - & - & - \\
      VideoTrack \cite{VideoTrack} & 72.9 & 81.9 & 69.8 & 70.2 & - & 76.4 & 83.8 & 88.7 & 83.1 & - & - & - \\
      ARTrack$_{256}$ \cite{ARTrack} & 73.5 & 82.2 & 70.9 & 70.4 & 79.5 & 76.6 & 84.2 & 88.7 & 83.5 & 46.4 & 56.5 & 52.3 \\
      ARTrack$_{384}$ \cite{ARTrack} & 75.5 & 84.3 & 74.3 & 72.6 & 81.7 & 79.1 & \color{blue}85.1 & 89.1 & 84.8 & 51.9 & 62.0 & 58.5 \\
      \hline
      \textbf{{\tracker}$_{256}$} & 73.0 & 82.9 & 68.6 & 69.7 & 79.9 & 76.6 & 84.7 & 89.7 & 83.7 & 47.5 & 57.8 & 53.5 \\
      \textbf{{\tracker}$_{384}$} & \color{blue}77.0 & \color{red}87.9 & \color{blue}75.1 & \color{blue}73.2 & \color{blue}83.2 & \color{blue}80.6 & \color{blue}85.1 & \color{blue}90.1 & \color{blue}84.9 & \color{blue}52.4 & \color{blue}63.9 & \color{blue}60.1 \\
      \textbf{{\tracker}-L$_{384}$} & \color{red}78.2& \color{blue}87.2 & \color{red}77.3 & \color{red}74.0 & \color{red}84.2 & \color{red}82.3 & \color{red}86.1 & \color{red}91.0 & \color{red}86.7 & \color{red}53.9 & \color{red}65.4 & \color{red}61.7 \\
    \bottomrule
    \end{tabular} }
\end{table*}

\begin{table*}[t]
\centering
\caption{Comparison with SOTA methods on TNL2K \cite{tnl2k} and OTB100 \cite{OTB2015} in AUC score. The best two results are highlighted in {\color{red}red} and {\color{blue}blue}, respectively.}
\label{tab:tnl2k}
\resizebox{\textwidth}{!}{
\begin{tabular}{c|cccccccccc|ccc}
\toprule
& \tabincell{c}{ATOM \\ \cite{ATOM}} & \tabincell{c}{Ocean \\ \cite{Ocean}} & \tabincell{c}{DiMP \\ \cite{DiMP50}} & \tabincell{c}{TransT \\ \cite{transt}} & \tabincell{c}{TransInMo \\ \cite{TransInMo}} & \tabincell{c}{OSTrack \\ \cite{ostrack}} & \tabincell{c}{SBT \\ \cite{SBT}} & \tabincell{c}{Mixformer \\ \cite{mixformer}} & \tabincell{c}{SeqTrack \\ \cite{seqtrack}} & \tabincell{c}{ARTrack \\ \cite{ARTrack}} & \textbf{{\tracker}$_{256}$} & \textbf{{\tracker}$_{384}$} & \textbf{{\tracker}-L$_{384}$} \\
\midrule
TNL2K & 40.1 & 38.4 & 44.7 & 50.7 & 52.0 & 55.9 & - & - & 56.4 & 59.8 & 58.0 & \color{blue}60.9 & \color{red}61.7 \\
OTB100 & 66.3 & 68.4 & 68.4 & 69.6 & 71.1 & - & 70.9 & 70.0 & - & - & \color{blue}72.3 & \color{blue}72.3 & \color{red}72.4 \\
\bottomrule
\end{tabular} }
\end{table*}

\begin{table*}[t]
\centering
\caption{State-of-the-art comparison on VOT2020 \cite{VOT2020}. The best two results are highlighted in {\color{red}red} and {\color{blue}blue}, respectively.}
\resizebox{\textwidth}{!}{
\begin{tabular}{l|cccccccccc|ccc}
\toprule
& \tabincell{c}{STM \\ \cite{STM}}  & \tabincell{c}{SiamMask \\ \cite{SiamMask}}  & \tabincell{c}{Ocean \\ \cite{Ocean}}  & \tabincell{c}{D3S \\ \cite{D3s}} & \tabincell{c}{AlphaRef \\ \cite{Alpha-Refine}} & \tabincell{c}{Ocean+ \\ \cite{ocean+}} & \tabincell{c}{STARK \\ \cite{stark}} & \tabincell{c}{SBT \\ \cite{SBT}}  & \tabincell{c}{Mixformer \\ \cite{mixformer}}  &  \tabincell{c}{SeqTrack \\ \cite{seqtrack}} & \textbf{{\tracker}$_{256}$} & \textbf{{\tracker}$_{384}$} & \textbf{{\tracker}-L$_{384}$}  \\
\midrule
EAO$(\uparrow)$ & 0.308 & 0.321 & 0.430 & 0.439 & 0.482 & 0.491 & 0.505 & 0.515 & 0.535 & 0.522 & 0.555 & {\color{blue}0.581} & {\color{red}0.605} \\
Accuracy$(\uparrow)$ & 0.751 & 0.624 & 0.693 & 0.699 & 0.754 & 0.685 & 0.759 & 0.752 & 0.761 & - & 0.751 & {\color{red}0.764} & {\color{blue}0.761} \\
Robustness$(\uparrow)$ & 0.574 & 0.648 & 0.754 & 0.769 & 0.777 & 0.842 & 0.819 & 0.825 & 0.854 & - & 0.864 & {\color{blue}0.877} & {\color{red}0.902} \\
\bottomrule
\end{tabular} }
\label{tab:vot2020}
\end{table*}

\textbf{One-shot Training.}
If a single neural network model can perform inference across multiple tasks simultaneously, it presents significant advantages. This not only reduces the need to hand craft models with appropriate inductive biases for each domain but also increases the quantity and diversity of available training data.

For RGB tracking task, similar to other work \cite{stark,ostrack,mixformer}, we use training datasets containing LaSOT \cite{LaSOT}, GOT-10k \cite{got10k}, TrackingNet \cite{TrackingNet}, and COCO \cite{COCO} to train our fundamental video-level tracking model.
In terms of input data, we take the video sequence including three reference frames with $192 \times 192$ pixels and two search frames with $384 \times 384$ pixels as the input to the model.

For multi-modal tracking tasks, in contrast to tracking algorithms \cite{TBSI,depthtrack,CEUTrack,vipt} trained independently on single downstream datasets, our goal is to jointly train multiple tracking tasks (i.e., RGB-T tracking, RGB-D tracking, and RGB-E tracking) at once.
We train our universal modal-awareness tracking model in a one-shot manner on joint thermal infrared (i.e., LasHeR \cite{lasher}, which aligns RGB and infrared data), depth (i.e., DepthTrack \cite{depthtrack}, which aligns RGB and depth data), and event (i.e., VisEvent \cite{visevent}, which aligns RGB and event data) datasets, supervising its predicted bounding boxes with the same loss function.

Specifically, we adopt the focal loss\cite{focalloss} as classification loss $L_{cls}$, and the $L_1$ loss and $GIoU$ loss\cite{GIoU} as regression loss. The total loss $L$ can be formulated as:
   \begin{equation}
      L_{total} = L_{cls} + \lambda_{1}L_{1} + \lambda_{2}L_{GIoU},
     \label{eq:loss}
   \end{equation}
where $\lambda_{1}$ = 5 and $\lambda_{2}$ = 2 are the regularization parameters. Since we use video segments for modeling, the task loss is computed independently for each video frame, and the final loss is averaged over the length of the search frames.

\textbf{Universal Inference.}
Algorithm \ref{algo:UM_ODTrack} summarizes the inference process of our models.
For RGB tracking, we follow the same tracking procedure as other transformer trackers \cite{ostrack,mixformer}.
Thanks to our one-shot training scheme and the gated modal-scalable perceiver module, for RGB-D, RGB-T, and RGB-E tracking tasks, we seamlessly perform inference for any tracking task using the same set of model parameters, without the need for additional multi-shot fine-tuning techniques.
In terms of input data, to align with the training setting, we incorporate three reference frames at equal intervals into our tracker during the inference phase. Concurrently, search frames and temporal token vectors are input frame-by-frame.

\begin{table*}[t]
\centering
\caption{Comparison with state-of-the-art methods on DepthTrack \cite{depthtrack}. The best two results are highlighted in {\color{red}red} and {\color{blue}blue}, respectively.}
\resizebox{\textwidth}{!}{
\begin{tabular}{l|ccccccccccc|cc}
\toprule
& \tabincell{c}{CA3DMS \\ \cite{CA3DMS}}  & \tabincell{c}{SiamM\_Ds \\ \cite{VOT2019}}  & \tabincell{c}{LTDSEd \\ \cite{VOT2019}}  & \tabincell{c}{LTMU\_B \\ \cite{VOT2020}} & \tabincell{c}{DDiMP \\ \cite{VOT2020}} & \tabincell{c}{DeT \\ \cite{depthtrack}}  & \tabincell{c}{OSTrack \\ \cite{ostrack}}  & \tabincell{c}{ProTrack \\ \cite{protrack}}  & \tabincell{c}{SPT \\ \cite{RGBD1K}}  &  \tabincell{c}{ViPT \\ \cite{vipt}} & \tabincell{c}{Un-Track \\ \cite{UnTrack}}  & \textbf{{\modaltracker}$_{256}$} & \textbf{{\modaltracker}$_{384}$}  \\
\midrule
F-score$(\uparrow)$ & 0.223 & 0.336 & 0.405 & 0.460 & 0.506 & 0.532 & 0.567 & 0.541 & 0.538 & 0.594 & 0.610 & {\color{blue}0.612} & {\color{red}0.693} \\
Re$(\uparrow)$ & 0.228 & 0.264 & 0.382 & 0.417 & 0.475 & 0.506 & 0.563 & 0.535 & 0.549 & 0.596 & 0.608 & {\color{blue}0.622} & {\color{red}0.707} \\
Pr$(\uparrow)$ & 0.218 & 0.463 & 0.430 & 0.512 & 0.540 & 0.560 & 0.572 & 0.547 & 0.527 & 0.592 & {\color{blue}0.611} & 0.603 & {\color{red}0.678} \\
\bottomrule
\end{tabular} }
\label{tab:depthtrack}
\end{table*}

\begin{table*}[t]
\centering
\caption{Comparison with state-of-the-art methods on VOT-RGBD2022 \cite{VOT2022}. The best two results are highlighted in {\color{red}red} and {\color{blue}blue}, respectively.}
\resizebox{\textwidth}{!}{
\begin{tabular}{l|ccccccccccc|cc}
\toprule
& \tabincell{c}{ATOM \\ \cite{ATOM}}  & \tabincell{c}{DiMP \\ \cite{DiMP50}}  & \tabincell{c}{STARK\_RGBD \\ \cite{VOT2021}} & \tabincell{c}{KeepTrack \\ \cite{VOT2022}} & \tabincell{c}{DeT \\ \cite{depthtrack}} & \tabincell{c}{OSTrack \\ \cite{ostrack}} & \tabincell{c}{SBT\_RGBD \\ \cite{SBT}} & \tabincell{c}{ProTrack \\ \cite{protrack}}  & \tabincell{c}{SPT \\ \cite{RGBD1K}}  &  \tabincell{c}{ViPT \\ \cite{vipt}} & \tabincell{c}{Un-Track \\ \cite{UnTrack}}  & \textbf{{\modaltracker}$_{256}$} & \textbf{{\modaltracker}$_{384}$}  \\
\midrule
EAO$(\uparrow)$ & 0.505 & 0.543 & 0.647 & 0.606 & 0.657 & 0.676 & 0.708 & 0.651 & 0.651 & 0.721 & 0.721 & {\color{blue}0.780} & {\color{red}0.790} \\
Accuracy$(\uparrow)$ & 0.698 & 0.703 & 0.803 & 0.753 & 0.760 & 0.803 & 0.809 & 0.801 & 0.798 & 0.815 & {\color{blue}0.820} & 0.814 & {\color{red}0.825} \\
Robustness$(\uparrow)$ & 0.688 & 0.731 & 0.798 & 0.797 & 0.845 & 0.833 & 0.864 & 0.802 & 0.851 & 0.871 & 0.869 & {\color{blue}0.948} & {\color{red}0.949} \\
\bottomrule
\end{tabular} }
\label{tab:vot_rgbd22}
\end{table*}

\begin{table*}[t]
\centering
\caption{Comparison with state-of-the-art methods on RGBT234 \cite{rgbt234}. The best two results are highlighted in {\color{red}red} and {\color{blue}blue}, respectively.}
\resizebox{\textwidth}{!}{
\begin{tabular}{l|cccccccccc|cc}
\toprule
& \tabincell{c}{MANet \\ \cite{MANet}} & \tabincell{c}{CAT \\ \cite{CAT}} & \tabincell{c}{HMFT \\ \cite{HMFT}} & \tabincell{c}{APFNet \\ \cite{APFNet}}  & \tabincell{c}{DMCNet \\ \cite{DMCNet}}  & \tabincell{c}{ProTrack \\ \cite{protrack}} & \tabincell{c}{ViPT \\ \cite{vipt}} & \tabincell{c}{TBSI \\ \cite{TBSI}} & \tabincell{c}{BAT \\ \cite{BAT}} & \tabincell{c}{Un-Track \\ \cite{UnTrack}} & \textbf{{\modaltracker}$_{256}$} & \textbf{{\modaltracker}$_{384}$}  \\
\midrule
MSR$(\uparrow)$ & 0.539 & 0.561 & 0.568 & 0.579 & 0.593 & 0.587 & 0.617 & 0.638 & 0.641 & 0.625 & {\color{blue}0.692} & {\color{red}0.708} \\
MPR$(\uparrow)$ & 0.777 & 0.804 & 0.788 & 0.827 & 0.839 & 0.786 & 0.835 & 0.871 & 0.868 & 0.842 & {\color{blue}0.915} & {\color{red}0.938} \\
\bottomrule
\end{tabular} }
\label{tab:rgbt234}
\end{table*}

\begin{figure}[t]
\begin{center}
\subfloat[Success Plot]{
\includegraphics[width=0.48\linewidth]{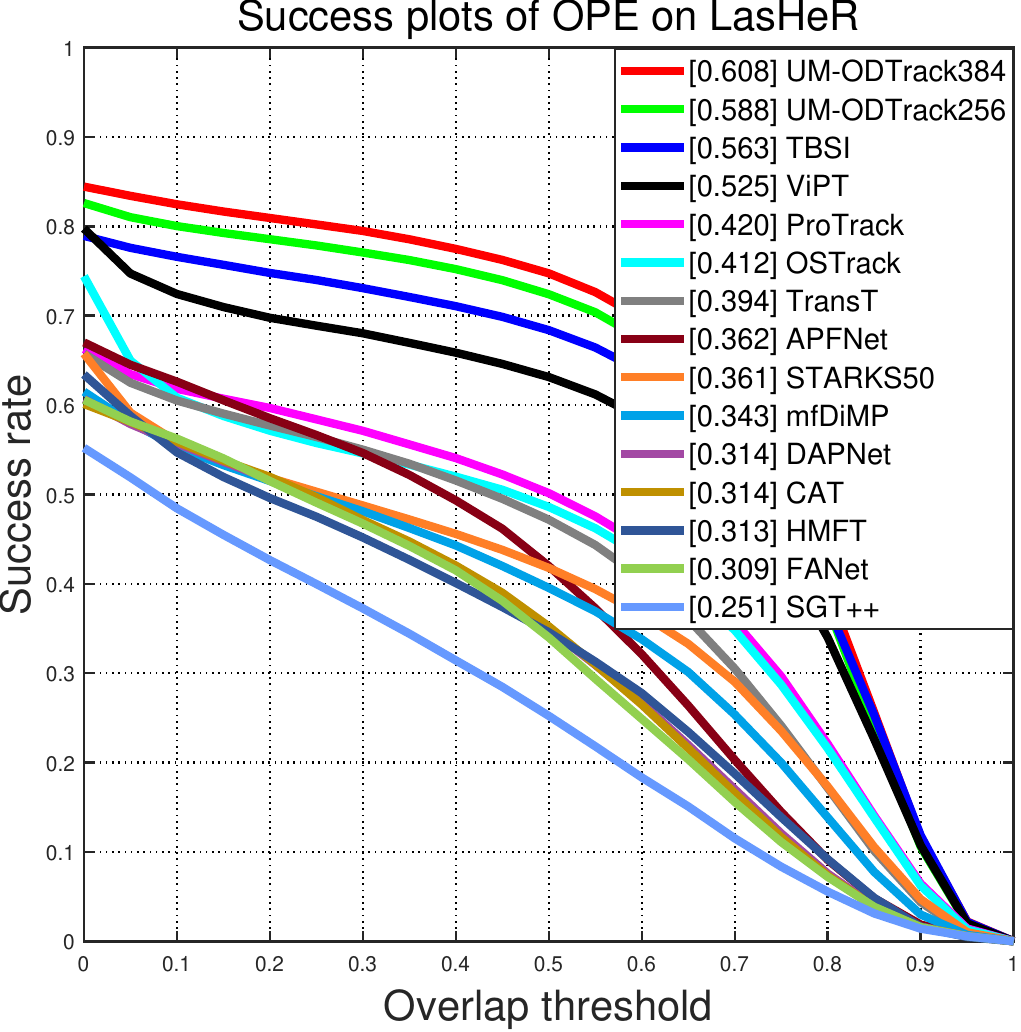}}
\subfloat[Precision Plot]{
\includegraphics[width=0.48\linewidth]{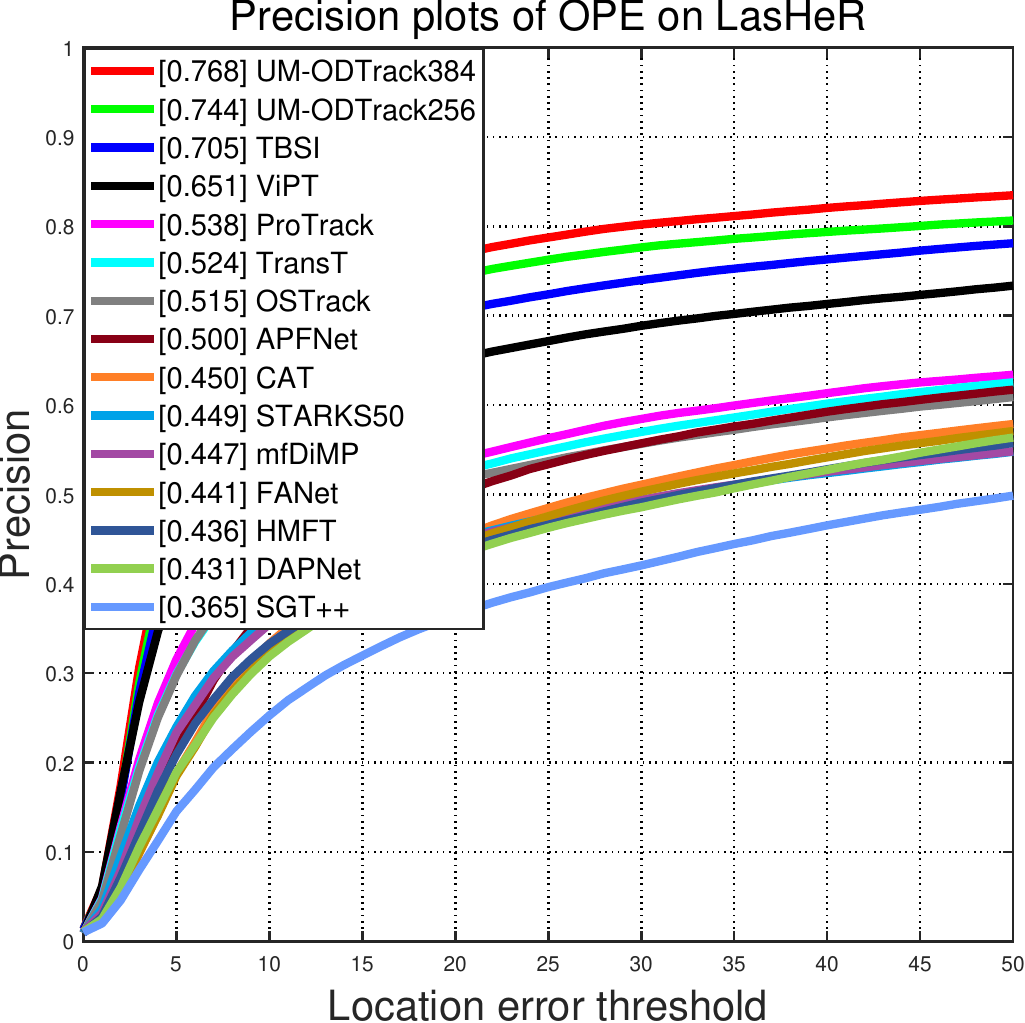}}
\end{center}
\caption{Success and precision plots on LasHeR \cite{lasher}.}
\label{fig:lasher}
\end{figure}

\begin{figure}[t]
\begin{center}
\subfloat[Success Plot]{
\includegraphics[width=0.48\linewidth]{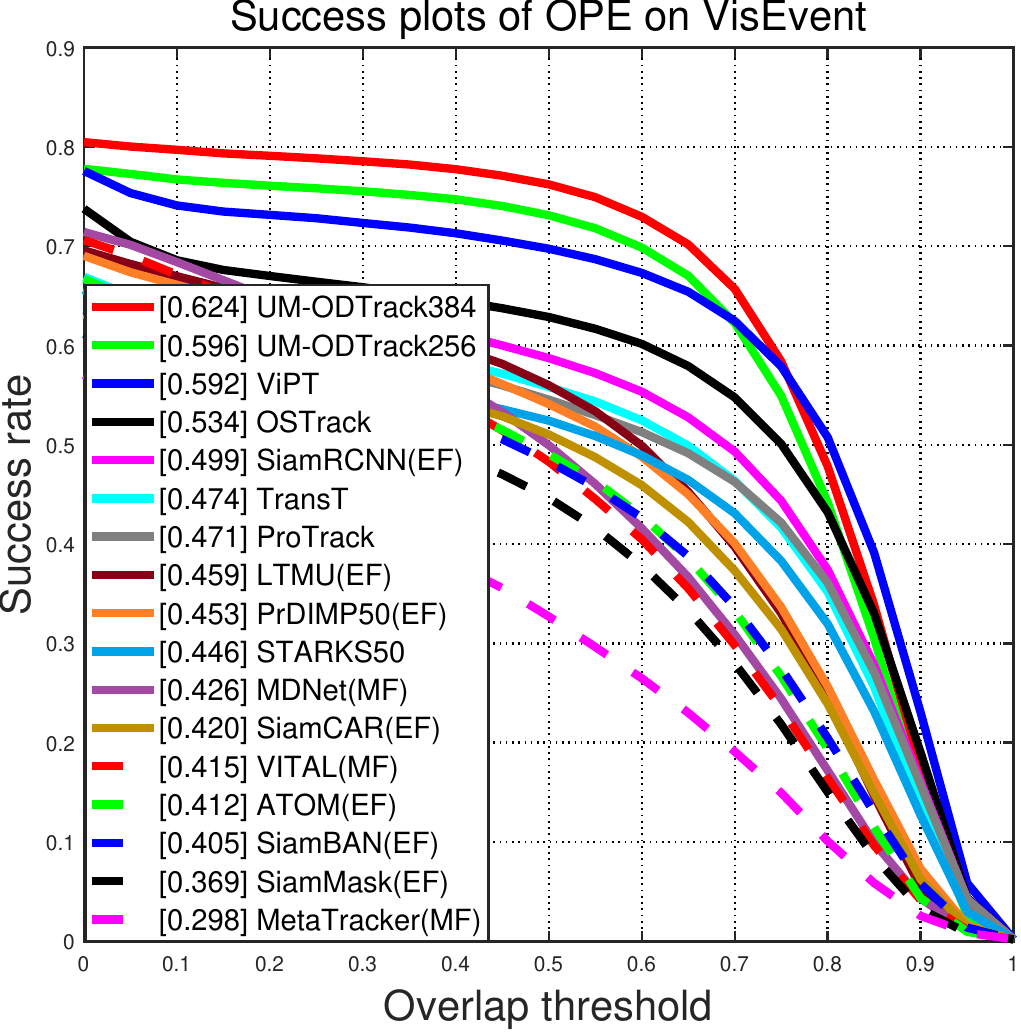}}
\subfloat[Precision Plot]{
\includegraphics[width=0.48\linewidth]{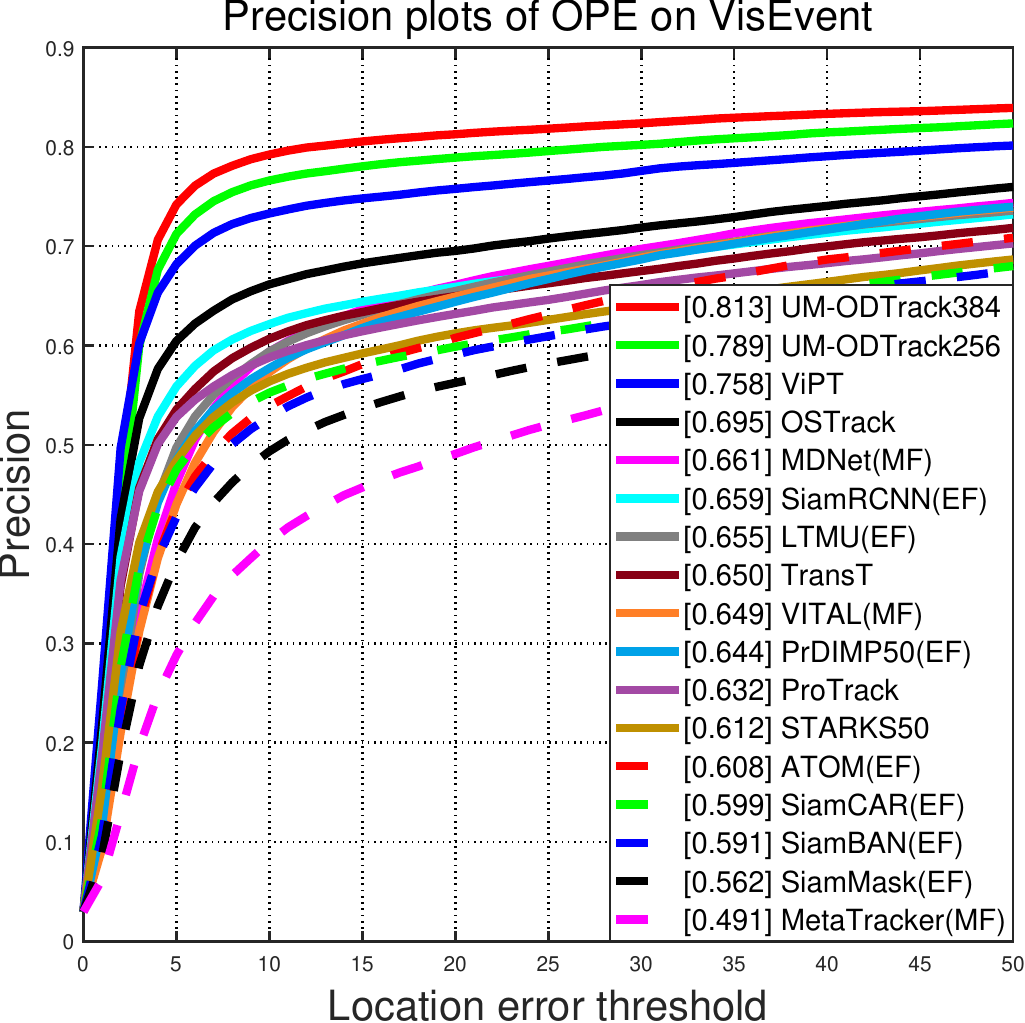}}
\end{center}
\caption{Success and precision plots on VisEvent \cite{visevent}.}
\label{fig:visevent}
\end{figure}

\section{Experiments}
    
\subsection{Implementation Details}
   We use ViT-Base \cite{vit} model as the visual encoder, and its parameters are initialized with MAE\cite{MAE} pre-training parameters.
   We employ the AdamW to optimize the network parameters with initial learning rate of $1 \times 10^{-5}$ for the backbone, $1 \times 10^{-4}$ for the rest, and set the weight decay to $10^{-4}$.
   $60,000$ image pairs are randomly sampled in each epoch.
   For the RGB tracking task, we set the training epochs to 300 epochs. 
   The learning rate drops by a factor of 10 after 240 epochs.
   For the multi-modal tracking tasks, we set the training epochs to 15 epochs. 
   The learning rate drops by a factor of 10 after 10 epochs.
   The model is conducted on a server with two 80GB Tesla A100 GPUs and set the batch size to be 8.

\subsection{Comparison with the SOTA}

We compare our {\tracker} and {\modaltracker} with the state-of-the-art trackers on seven visible (including LaSOT \cite{LaSOT}, TrackingNet \cite{TrackingNet}, GOT10K \cite{got10k}, LaSOT$_{\rm{ext}}$ \cite{lasot-ext}, VOT2020 \cite{VOT2020}, TNL2K \cite{tnl2k}, and OTB100 \cite{OTB2015}) and five multi-modal tracking benchmarks (including LasHeR \cite{lasher}, RGBT234 \cite{rgbt234}, DepthTrack \cite{depthtrack}, VOT-RGBD2022 \cite{VOT2022}, and VisEvent \cite{visevent}). 
Our {\tracker} and {\modaltracker} demonstrate excellent performance on these datasets.

\textbf{GOT10K.}
GOT10K is a large-scale tracking dataset that contains more than 10,000 video sequences. The GOT10K benchmark proposes a protocol, which the trackers only use its training set for training. We follow the protocol to train our framework.
The results are recorded in Tab.\ref{tab:results}. Among previous methods, ARTrack$_{384}$ \cite{ARTrack} without a video-level sampling strategy achieves a sota performance on the AO (average overlap), SR$_{0.5}$, and SR$_{0.75}$ (success rates at threshold 0.5 and 0.75) metrics, respectively.
Benefiting from the proposed new video-level sampling strategy, our {\tracker}$_{384}$ obtains a new state-of-the-art that scores 77.0\%, 87.9\%, and 75.1\% on the AO, SR$_{0.5}$, and SR$_{0.75}$ metrics, respectively.
The results demonstrate that one benefit of our {\tracker} comes from the video-level sampling strategy, which is design to release the potential of our model.

\begin{table}[t]
\caption{Ablation study of different association methods on LaSOT \cite{LaSOT}.}
\centering
\setlength{\tabcolsep}{5.5mm}{
\begin{tabular}{c|ccc}
\toprule
Method & AUC & $P_{Norm}$ & $P$ \\
\midrule
\textit{Baseline} & 70.1 & 80.2 & 76.9 \\
$w/o$ \textit{Token} & 71.0 & 81.1 & 78.0 \\
\textit{Separate} & 72.2 & 82.3 & 79.2 \\
\textit{Concatenation} & \textbf{72.8} & \textbf{83.0} & \textbf{80.3} \\
\bottomrule
\end{tabular} }
\label{tab:prop_method}
\end{table}

\begin{table}[t]
\caption{Ablation study of search video-clip length on LaSOT \cite{LaSOT}.}
\centering
\setlength{\tabcolsep}{5mm}{
\begin{tabular}{c|ccc}
\toprule
Sequence Length & AUC & $P_{Norm}$ & $P$ \\
\midrule
2 & 72.8 & 83.0 & 80.3 \\
3 & \textbf{73.1} & \textbf{83.0} & \textbf{80.4} \\
4 & 72.5 & 82.9 & 79.9 \\
5 & 72.0 & 82.1 & 79.3 \\
\bottomrule
\end{tabular}  }
\label{tab:seq_len}
\end{table}

\textbf{LaSOT.}
LaSOT is a large-scale long-term tracking benchmark that includes 1120 sequences for training and 280 sequences for testing. 
As shown in Tab.\ref{tab:results}, it can be seen that our {\tracker}$_{384}$ achieves good tracking results with an interesting temporal token attention mechanism. Compared with the performance from the latest ARTrack, our {\tracker}$_{384}$ achieves 0.6\%, 1.5\%, and 1.5\% gains in terms of AUC, P${_{\rm{Norm}}}$ and P score, respectively. 
The results show that the spatio-temporal features with target association dependencies that have been learned by the tracker can provide reliable target localization.
Furthermore, since our temporal token is designed to associate target instance to improve robustness and accuracy under multiple tracking challenges, i.e., fast motion, backgroud clutter, viewpoint change, and scale variation, etc. Thus, as depicted in the Fig.\ref{fig:attrs}, we present an attribute evaluation from the LaSOT dataset to illustrate how our token association mechanism aids our tracker in learning spatio-temporal trajectory information about the target instance, significantly enhancing target localization in long-term tracking scenarios.

\textbf{TrackingNet.}
TrackingNet is a large-scale short-term dataset that provides a test set with 511 video sequences.
As reported in Tab.\ref{tab:results}, by implementing cross-frame association of target instance, our {\tracker}$_{384}$ gets a success score (AUC) of 85.1\%, a normalized precision score (P$_{Norm}$) of 90.1\%, and a precision score (P) of 84.9\%, outperforming previous high-preformance tracker SeqTrack without token association by 1.2\%, 1.3\% and 1.3\%, respectively. 
Meanwhile, compared to the recent video-level tracker VideoTrack \cite{VideoTrack} without temporal token association, our {\tracker} outperforms the AUC, P$_{Norm}$, and P metrics by 1.3\%, 1.4\% and 1.8\% scores respectively.
This indicates that our temporal token can effectively associate target object across search frames, and this novel association way can enhance the generalization capability of our {\tracker} across multiple tracking scenarios.

\begin{table}[t]
\caption{Ablation study of different sampling ranges on LaSOT benchmark \cite{LaSOT}.}
\centering
\setlength{\tabcolsep}{5mm}{
\begin{tabular}{c|ccc}
\toprule
Sample Range & AUC & $P_{Norm}$ & $P$ \\
\midrule
200 & 72.8 & 83.0 & 80.3 \\
400  & \textbf{73.2} & \textbf{83.2} & \textbf{80.6} \\
800  & 73.0 & 83.3 & 80.4 \\
1200  & 73.0 & 83.1 & 80.1 \\
\bottomrule
\end{tabular} }
\label{tab:sampling_range}
\end{table}

\begin{table}[t]
    \centering
    \caption{Ablation study on different gated perceivers. Results are reported in percentage (\%).}
    \label{tab:modules}
    \resizebox{\linewidth}{!}{
    \begin{tabular}{c|c|ccc|cc|cc}
    \toprule
    \multicolumn{1}{c|}{\multirow{2}{*}{\#}} & \multicolumn{1}{c|}{\multirow{2}{*}{Method}} & \multicolumn{3}{c|}{DepthTrack} & \multicolumn{2}{c|}{LasHeR} & \multicolumn{2}{c}{VisEvent} \\
     \cline{3-9}
     & & F-score($\uparrow$) & Re($\uparrow$) & Pr($\uparrow$) & SR($\uparrow$) & PR($\uparrow$) & SR($\uparrow$) & PR($\uparrow$) \\
     \midrule
     1 & \textit{Baseline} & 66.7 & 68.0 & 65.5 & 58.4 & 73.7 & 59.6 & 80.0 \\
     2 & \textit{+ Conditional gate} & 68.0 & 69.5 & 66.6 & 59.4 & 75.3 & 60.6 & 78.9 \\
     3 & \textit{+ GMP} & \textbf{69.3} & \textbf{70.7} & \textbf{67.8} & \textbf{60.8} & \textbf{76.8} & \textbf{62.4} & \textbf{81.3} \\
    \bottomrule
    \end{tabular} }
\end{table}

\textbf{LaSOT$_{\rm{ext}}$.}
LaSOT$_{\rm{ext}}$ is the extended version of LaSOT, which comprises 150 long-term video sequences.
As reported in Tab.\ref{tab:results}, our method achieves the good tracking results that outperform most compared trackers. For example, our tracker gets a AUC of 52.4\%, $P_{Norm}$ score of 63.9\%, and $P$ score of 60.1\%, outperforming the ARTrack by 0.5\%, 1.9\%, and 1.6\%, respectively.
Furthermore, our {\tracker} also outperforms the advanced image-pair matching based trackers OSTrack by 1.9\% in terms of success score.
The results meet our expectation that video-level modeling has more stable object localization capabilities in complex long-term tracking scenarios.

\textbf{VOT2020.}
VOT2020\cite{VOT2020} contains 60 challenging sequences, and it uses binary segmentation masks as the groundtruth. We use Alpha-Refine \cite{Alpha-Refine} as a post-processing network for {\tracker} to predict segmentation masks. The Expected Average Overlap (EAO) metric is used to evaluate the proposed tracker and other advanced trackers.
As shown in Tab.\ref{tab:vot2020}, our {\tracker}$_{384}$ and -L$_{384}$ achieve the best results with EAO of 58.1\% and 60.5\% on mask evaluations.
In terms of EAO metrics, {\tracker} outperforms by 6.6\% and 9\% scores, respectively, compared to trackers that do not explore temporal relations (i.e., SBT \cite{SBT} and Ocean+ \cite{ocean+}). These results show that by injecting a temporal token attention, our {\tracker} is robust in complex tracking scenarios.

\textbf{TNL2K and OTB100.}
We evaluate our tracker on TNL2K\cite{tnl2k} and OTB100\cite{OTB2015} benchmarks. They include 700 and 100 video sequences, respectively. These results in Tab.\ref{tab:tnl2k} show that the {\tracker}$_{384}$ and -L$_{384}$ achieve the best performance on TNL2K and OTB100 benchmarks. For example, 
our {\tracker}$_{384}$ obtains AUC scores of 60.9\% and 72.3\% on the TNL2K and OTB100 datasets, respectively. Compared to ARTrack on the TNL2K dataset, {\tracker} outperforms it by 1.1\%. 
Meanwhile, compared to the non-autoregressive tracker Mixformer, our {\tracker} has a 2.3\% higher AUC score on the OTB100 dataset.
It can be observed that by employing an intriguing auto-regressive modeling approach to capture temporal context, our {\tracker} is capable of reducing model complexity and improving performance.

   \begin{figure}[t]
      \centering
      \includegraphics[width=1\linewidth]{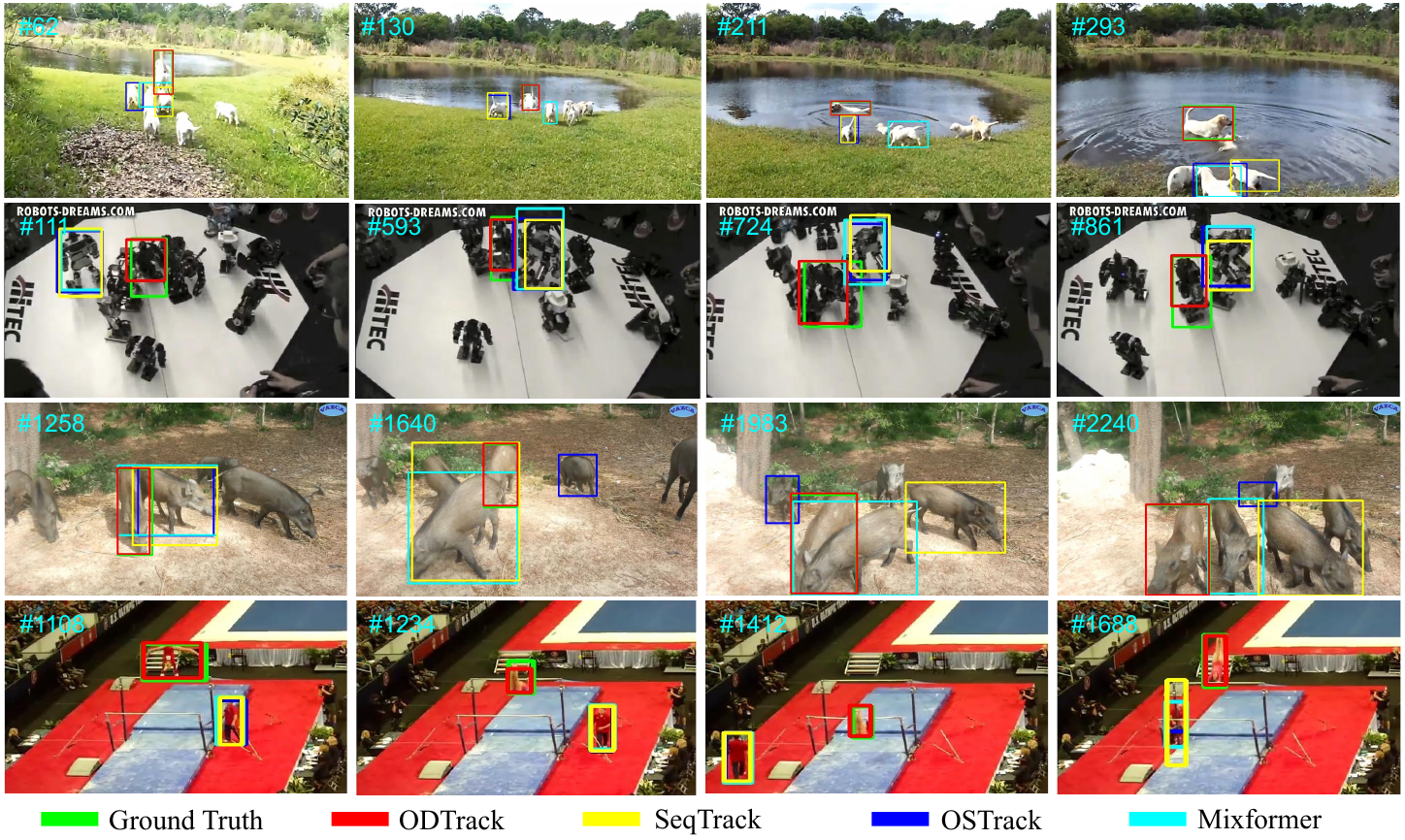}
       \caption{Qualitative comparison results of our tracker with other three SOTA trackers on LaSOT benchmark.
       }
       \label{fig:visual}
    \end{figure}

   \begin{figure}[t]
      \centering
      \includegraphics[width=1\linewidth]{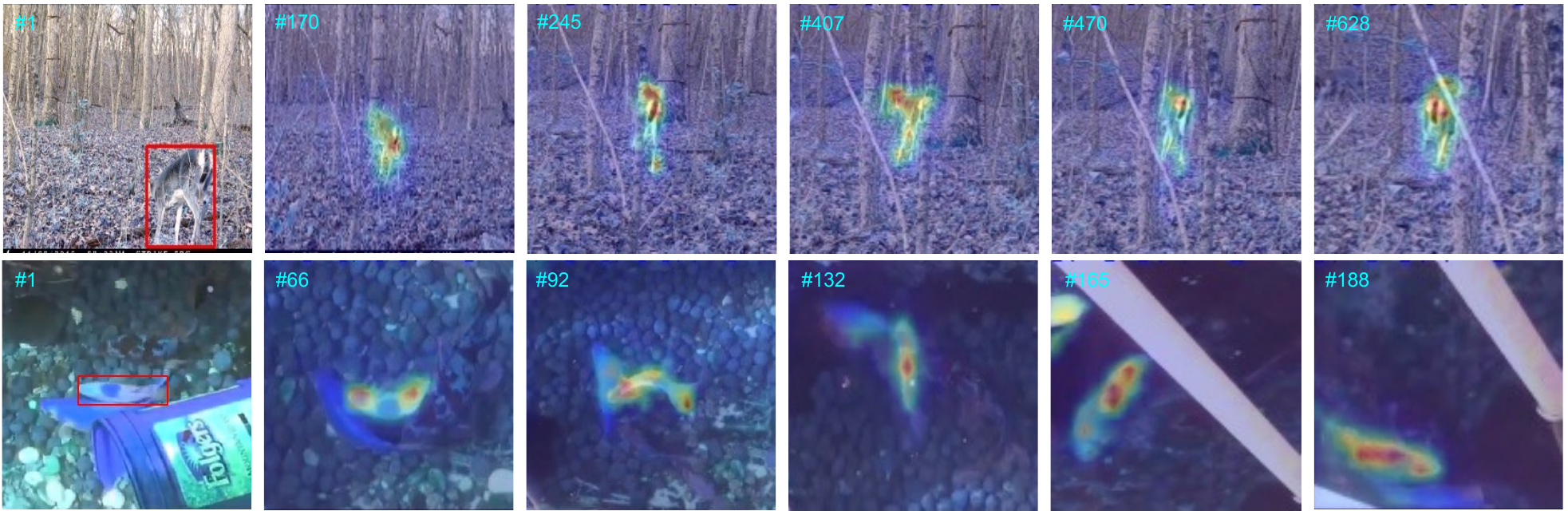}
       \caption{The attention maps of temporal token attention operation.}
       \label{fig:attnmap}
    \end{figure}

\textbf{DepthTrack.}
DepthTrack \cite{depthtrack} consists of 150 training and 50 test RGB-D long-term video sequences.
As shown in Tab.\ref{tab:depthtrack}, we compare our model with existing SOTA RGB-D trackers on it.
With the same image resolution setting, our {\modaltracker}$_{256}$ outperforms ViPT \cite{vipt} by 1.1\%, 2.6\%, and 1.8\% in terms of tracking precision (Pr), recall (Re) and F-score, respectively. Furthermore, due to the proposed efficient gated attention mechanism, our {\modaltracker}$_{384}$ obtains the SOTA performance in the field of RGB-D tracking.
Notably, {\modaltracker} achieves a large improvement in performance when the input size is increased from 256 to 384. 
This demonstrates that our temporal association approach incorporating large input resolution is particularly important in multi-modal long-term tracking scenarios.

\textbf{VOT-RGBD2022.}
VOT-RGBD2022 \cite{VOT2022} is a short-term tracking dataset that contains 127 RGB-D video sequences.
As reported in Tab.\ref{tab:vot_rgbd22}, compared to most other tracking algorithms, our tracker achieves a new state-of-the-art result. Specifically, our {\modaltracker}$_{256}$ obtains scores of 78.0\%, 81.4\%, and 94.8\% in terms of EAO, accuracy, and robustness metrics, respectively.
Compared with the latest unified tracker Un-Track \cite{UnTrack}, our {\modaltracker}$_{256}$ achieves 5.9\% and 7.9\% gains in terms of expected average overlap (EAO) and robustness score, respectively.
This indicates that our unified modeling technique is more effective towards universal feature learning, and can provide a suitable and stable feature space for each modality, i.e., depth modality.

\textbf{LasHeR.}
LasHeR \cite{lasher} is a large-scale RGB-T tracking dataset containing 245 short-term test video sequences.
The results are reported in Fig.\ref{fig:lasher}, our {\modaltracker} achieves surprising results, dramatically outperforming the previous SOTA RGB-T tracking algorithms, which exceeds the second place by 4.5\% and 6.3\% in terms of success plot and precision plot, respectively.
These results are consistent with our expectation that feature learning based on gated attention mechanism can adaptively extract and fuse different modal features to improve multi-modal tracking performance. 
Meanwhile, to verify that our gated perceiver can effectively solve multiple challenges including occlusion (NO), partial occlusion (PO), total occlusion (TO), low illumination (LI), low resolution (LR), deformation (DEF), background clutter (BC), motion blur (MB), thermal crossover (TC), camera moving (CM), fast motion (FM), scale variation (SV), hyaline occlusion (HO), high illumination (HI), abrupt illumination variation (AIV), similar appearance (SA), aspect ratio change (ARC), out-of-view (OV) and frame lost (FL), we show the attribute evaluation results for the LasHeR dataset. As shown in Fig.\ref{fig:lasher_attr}, our {\modaltracker} performs well on each attribute. 
Thus, it can be indicates that our video-level multi-modal modeling scheme with gated perceivers can effectively unify and fuse the multi-modal features, so that our tracker to solve complex tracking scenarios well.

\begin{table}[t]
    \centering
    \caption{Ablation study on conditional gate layers. Results are reported in percentage (\%).}
    \label{tab:gatelayers}
    \resizebox{\linewidth}{!}{
    \begin{tabular}{c|c|ccc|cc|cc}
    \toprule
    \multicolumn{1}{c|}{\multirow{2}{*}{\#}} & \multicolumn{1}{c|}{\multirow{2}{*}{Layer}} & \multicolumn{3}{c|}{DepthTrack} & \multicolumn{2}{c|}{LasHeR} & \multicolumn{2}{c}{VisEvent} \\
     \cline{3-9}
     & & F-score($\uparrow$) & Re($\uparrow$) & Pr($\uparrow$) & SR($\uparrow$) & PR($\uparrow$) & SR($\uparrow$) & PR($\uparrow$) \\
     \midrule
     1 & \textit{3,7,11} & 67.5 & 69.0 & 66.1 & 59.5 & 75.8 & 60.8 & 80.4 \\
     2 & \textit{1,3,5,7,9,11} & 68.3 & 69.8 & 66.9 & 59.9 & 75.7 & 61.1 & 80.8 \\
     3 & \textit{0-11} & \textbf{69.3} & \textbf{70.7} & \textbf{67.8} & \textbf{60.8} & \textbf{76.8} & \textbf{62.4} & \textbf{81.3} \\
    \bottomrule
    \end{tabular} }
\end{table}

\begin{table}[t]
    \centering
    \caption{Ablation study on gated modal-scalable perceiver layers. Results are reported in percentage (\%).}
    \label{tab:perceiverlayers}
    \resizebox{\linewidth}{!}{
    \begin{tabular}{c|c|ccc|cc|cc}
    \toprule
    \multicolumn{1}{c|}{\multirow{2}{*}{\#}} & \multicolumn{1}{c|}{\multirow{2}{*}{Layer}} & \multicolumn{3}{c|}{DepthTrack} & \multicolumn{2}{c|}{LasHeR} & \multicolumn{2}{c}{VisEvent} \\
     \cline{3-9}
     & & F-score($\uparrow$) & Re($\uparrow$) & Pr($\uparrow$) & SR($\uparrow$) & PR($\uparrow$) & SR($\uparrow$) & PR($\uparrow$) \\
     \midrule
     1 & \textit{1} & 68.8 & 70.3 & 67.3 & 59.7 & 75.3 & 61.2 & 80.6 \\
     2 & \textit{3} & \textbf{69.3} & \textbf{70.7} & \textbf{67.8} & \textbf{60.8} & \textbf{76.8} & \textbf{62.4} & \textbf{81.3} \\
     3 & \textit{6} & 69.1 & 70.6 & 67.7 & 59.8 & 75.5 & 61.3 & 80.7 \\
    \bottomrule
    \end{tabular} }
\end{table}

\textbf{RGBT234.}
RGBT234 \cite{rgbt234} contains 234 RGB-T tracking videos with approximately 116.6K image pairs. 
As shown in Fig.\ref{tab:rgbt234}, our {\modaltracker}$_{256}$ obtains scores of 69.2\% and 91.5\% in terms of the SR and PR metrics, respectively. Compared with the high-preformance RGB-T expert tracker BAT \cite{BAT}, our method achieves good tracking results that outperform 5.1\% and 4.7\% in terms of success plot and precision plot, respectively. 
This means that our GMP module can efficiently aggregate target information from thermal infrared modality for robust multi-modal tracking.

\textbf{VisEvent.}
VisEvent \cite{visevent} is the largest RGB-E tracking benchmark containing 320 test videos.
Comparison results are shown in Fig.\ref{fig:visevent}. Our {\modaltracker}$_{384}$ achieves a new SOTA tracking results with success score of 62.4\% and 81.3\% on precision score, respectively.
It can be seen that our {\modaltracker} equipped with the gated modal-scalable perceiver (GMP) module also achieves accurate tracking in event scenarios. This is consistent with our intuition that GMP module can be easily extended to different modality tracking scenarios and effectively improve the representation of multi-modal features.

\begin{table}[t]
    \centering
    \caption{Ablation study on different gated activation functions. Results are reported in percentage (\%).}
    \label{tab:activation}
    \resizebox{\linewidth}{!}{
    \begin{tabular}{c|c|ccc|cc|cc}
    \toprule
    \multicolumn{1}{c|}{\multirow{2}{*}{\#}} & \multicolumn{1}{c|}{\multirow{2}{*}{Method}} & \multicolumn{3}{c|}{DepthTrack} & \multicolumn{2}{c|}{LasHeR} & \multicolumn{2}{c}{VisEvent} \\
     \cline{3-9}
     & & F-score($\uparrow$) & Re($\uparrow$) & Pr($\uparrow$) & SR($\uparrow$) & PR($\uparrow$) & SR($\uparrow$) & PR($\uparrow$) \\
     \midrule
     1 & \textit{ReLU} & 65.0 & 66.3 & 63.8 & 58.7 & 75.5 & 59.3 & 78.9 \\
     2 & \textit{Sigmoid} & 67.0 & 68.4 & 65.7 & 59.0 & 75.2 & 59.7 & 80.1 \\
     3 & \textit{Tanh} & \textbf{69.3} & \textbf{70.7} & \textbf{67.8} & \textbf{60.8} & \textbf{76.8} & \textbf{62.4} & \textbf{81.3} \\
    \bottomrule
    \end{tabular} }
\end{table}

\begin{table}[t]
    \centering
    \caption{Ablation studies on multi-modal tracking benchmarks, i.e., DepthTrack \cite{depthtrack}, LasHeR \cite{lasher}, and VisEvent \cite{visevent}. Results are reported in percentage (\%).}
    \label{tab:multimodal_study}
    \resizebox{\linewidth}{!}{
    \begin{tabular}{c|c|ccc|cc|cc}
    \toprule
    \multicolumn{1}{c|}{\multirow{2}{*}{\#}} & \multicolumn{1}{c|}{\multirow{2}{*}{Method}} & \multicolumn{3}{c|}{DepthTrack} & \multicolumn{2}{c|}{LasHeR} & \multicolumn{2}{c}{VisEvent} \\
     \cline{3-9}
     & & F-score($\uparrow$) & Re($\uparrow$) & Pr($\uparrow$) & SR($\uparrow$) & PR($\uparrow$) & SR($\uparrow$) & PR($\uparrow$) \\
     \midrule
     1 & \textit{RGB-only} & 64.3 & 65.4 & 63.2 & 54.5 & 69.6 & 59.7 & 78.8 \\
     2 & \textit{Not Share tokenizers} & 62.3 & 63.6 & 61.0 & 48.5 & 63.3 & 57.1 & 77.2 \\
     3 & \textit{Adapter tuning} & 65.9 & 67.2 & 64.6 & 59.2 & 75.0 & 60.4 & 79.8 \\
     4 & \textit{Single-task training} & 67.8 & 69.3 & 66.4 & 58.7 & 75.5 & 60.6 & 79.8 \\
     5 & \textit{{\modaltracker}} & \textbf{69.3} & \textbf{70.7} & \textbf{67.8} & \textbf{60.8} & \textbf{76.8} & \textbf{62.4} & \textbf{81.3} \\
    \bottomrule
    \end{tabular} }
\end{table}

\subsection{Ablation Study}

\textbf{The effectiveness of token association.}
To investigate the effect of token association in Eq.\ref{eq:propagate}, we perform experiments whether propagating temporal token in Tab.\ref{tab:prop_method}. $w/o$ \textit{Token} denotes the experiment employing video-level sampling strategy without token association.
From the second and third rows, it can be observed that the absence of the token association mechanism leads to a decrease in the AUC score by 1.2\%.
This result indicates that token association plays a crucial role in cross-frame target association.
Furthermore, we conduct experiments to validate the effectiveness of two proposed token association methods in the video-level tracking framework in Tab.\ref{tab:prop_method}. We can be observe that both the \textit{separate} and \textit{concatenation} methods achieve significant performance improvements, with the \textit{concatenation} method showing slightly better results. This demonstrates the effectiveness of both attention mechanisms.

   \begin{figure}[t]
      \centering
      \includegraphics[width=1\linewidth]{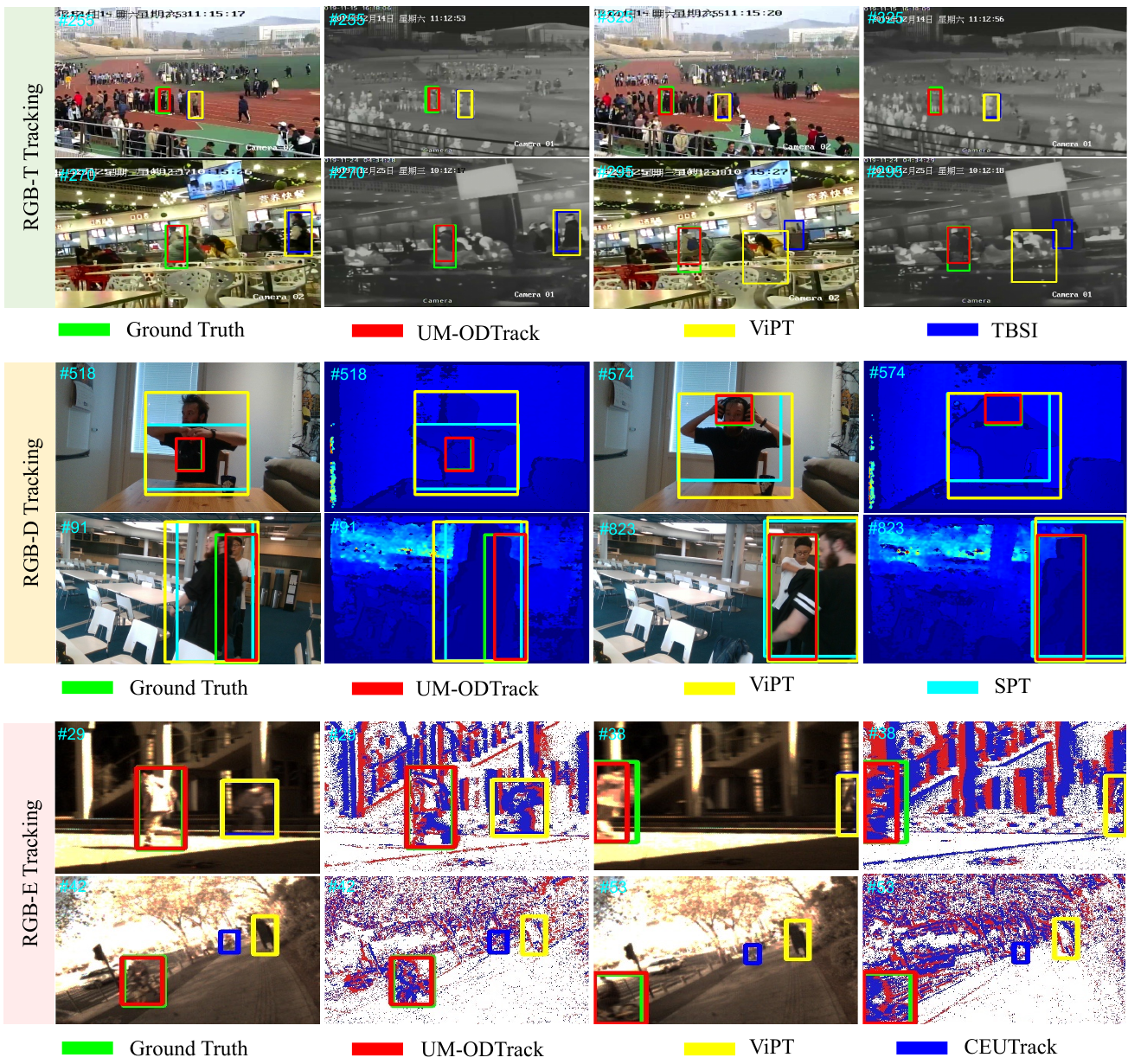}
       \caption{Qualitative comparison results of our tracker with other SOTA multi-modal trackers on LasHeR, DepthTrack, and VisEvent benchmark.
       }
       \label{fig:modal_visual}
    \end{figure}

\textbf{The length of search video-clip.}
As shown in Tab.\ref{tab:seq_len}, we ablate the impact of search video sequence length on the tracking performance. 
When the length of video clip increases from 2 to 3, the AUC metric improves by 0.3\%. However, continuous increment in sequence length does not lead to performance improvement, indicating that overly long search video clips impose a learning burden on the model. Hence, we should opt for an appropriate the length of search video clip.

Furthermore, to evaluate the impact of sequence length on multi-modal tracking performance, we conduct comparative experiments on the LasHeR \cite{lasher}, DepthTrack \cite{depthtrack}, and VisEvent \cite{visevent} benchmarks, as shown in Tab.\ref{tab:seq_length}. 
The choice of video sequence length is crucial for leveraging temporal information. When the sequence length increases from 2 to 3, our tracker achieves improvements in SR and F-score by 0.7\%, 0.2\%, and 1.6\% on the LasHeR, DepthTrack, and VisEvent benchmarks, respectively. These gains result from the effective modeling of object appearance changes and motion trajectories through multi-frame information. However, when the sequence length exceeds 3, the performance tends to plateau or slightly decline due to the accumulation of cross-modal temporal noise. This confirms that an appropriately selected sequence length can provide complementary information, while excessively long sequences are more likely to introduce redundant or noisy contextual signals. Therefore, our UM-ODTrack adopts a sequence length of 3 as an optimal setting to capture contextual information over a suitable temporal span.

\textbf{The sampling range.}
To validate the impact of sampling range on algorithm performance, we conduct experiments on the sampling range of video frames in Tab.\ref{tab:sampling_range}.
When the sampling range is expanded from 200 to 1200, there is a noticeable improvement in performance on the AUC metric, indicating that the video-level framework can learn target trajectory information from a larger sampling range.

\begin{figure}[t]
\begin{center}
\subfloat[Adapter/Prompt tuning]{
\includegraphics[width=0.36\linewidth]{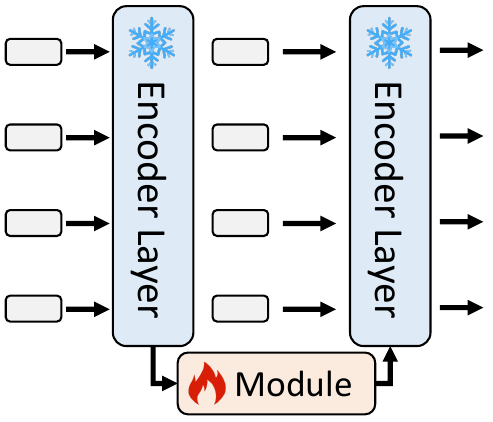}}
~~~~~~~~
\subfloat[Full tuning]{
\includegraphics[width=0.36\linewidth]{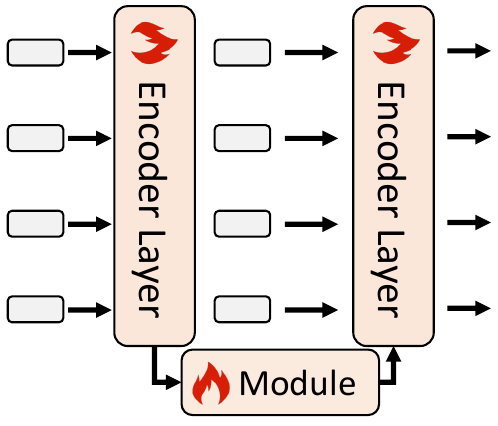}}
\end{center}
\caption{Adapter/Prompt fine-tuning and full fine-tuning. Snow means that the learning parameters are frozen. Fire represents that the parameters are learnable.}
\label{fig:adapter}
\end{figure}

\begin{table}[t]
    \centering
    \caption{Ablation study of modality weight ratios on LasHeR \cite{lasher}, DepthTrack \cite{depthtrack}, and VisEvent \cite{visevent}. The weight ratio follows the order depth: infrared: event = 1:1:1. Results are reported in percentage (\%).}
    \label{tab:weight_ratio}
    \resizebox{\linewidth}{!}{
    \begin{tabular}{c|c|ccc|cc|cc}
    \toprule
    \multicolumn{1}{c|}{\multirow{2}{*}{\#}} & \multicolumn{1}{c|}{\multirow{2}{*}{Ratio}} & \multicolumn{3}{c|}{DepthTrack} & \multicolumn{2}{c|}{LasHeR} & \multicolumn{2}{c}{VisEvent} \\
     \cline{3-9}
     & & F-score($\uparrow$) & Re($\uparrow$) & Pr($\uparrow$) & SR($\uparrow$) & PR($\uparrow$) & SR($\uparrow$) & PR($\uparrow$) \\
     \midrule
     1 & \textit{1:1:1} & \textbf{69.3} & \textbf{70.7} & \textbf{67.8} & \textbf{60.8} & \textbf{76.8} & \textbf{62.4} & 81.3 \\
     2 & \textit{2:1:1} & 69.1 & 70.3 & 67.2 & 60.4 & 76.6 & 61.7 & 80.6 \\
     3 & \textit{1:2:1} & 68.3 & 69.5 & 66.5 & 60.8 & 76.7 & 62.1 & 81.2 \\
     4 & \textit{1:1:2} & 68.2 & 68.8 & 65.9 & 60.2 & 76.3 & 62.3 & \textbf{81.5} \\
    \bottomrule
    \end{tabular} }
\end{table}

\textbf{The effectiveness of gated perceivers and gated activation functions.}
We conduct experiments to validate the effectiveness of the two proposed components, the \textit{conditional gate} and the \textit{gated modal-scalabel perceiver} (GMP), under the universal modal-awareness tracking framework, as shown in Tab.\ref{tab:modules}.
The baseline method refers to the dual-stream version of ODTrack.
By adding the conditional gate module to the baseline, our tracker's performance is improved on three downstream tracking datasets. For example, the tracker equipped with conditional gate achieved a 1.3\% improvement in the \textit{F-score} metric on the DepthTrack \cite{depthtrack} benchmark.
Furthermore, by incorporating GMP into our model, its performance is further improved.
This demonstrates the effectiveness of our proposed two gated modules.
Furthermore, to investigate the effect of gated activation functions, we perform experiments with different gated activation functions in Tab.\ref{tab:activation}. 
Compared to \textit{ReLU} and \textit{sigmoid}, the \textit{tanh} activation function performs the best. This result suggests that in our gated perceivers, the \textit{tanh} gated function is more suitable for learning and representing general multi-modal tracking tasks, potentially offering better generalization capabilities.

\textbf{The number of conditional gate and GMP layers.}
We compare the impact of the number of layers in the conditional gate and GMP (Gated Modal-scalable Perceiver) on the model, respectively. The experimental results are recorded in Tab.\ref{tab:gatelayers} and Tab.\ref{tab:perceiverlayers}.
For the conditional gate, with an increase in the number of layers, there is a corresponding improvement in the performance of our tracker. This implies that integrating more layers in the universal modal encoder contributes to learning multi-modal representations.
On the other hand, our {\modaltracker} achieves good tracking results when using three or six layers in the GMP module. To balance speed and performance, we chose to use a three-layer configuration.

   \begin{figure}[t]
      \centering
      \includegraphics[width=0.95\linewidth]{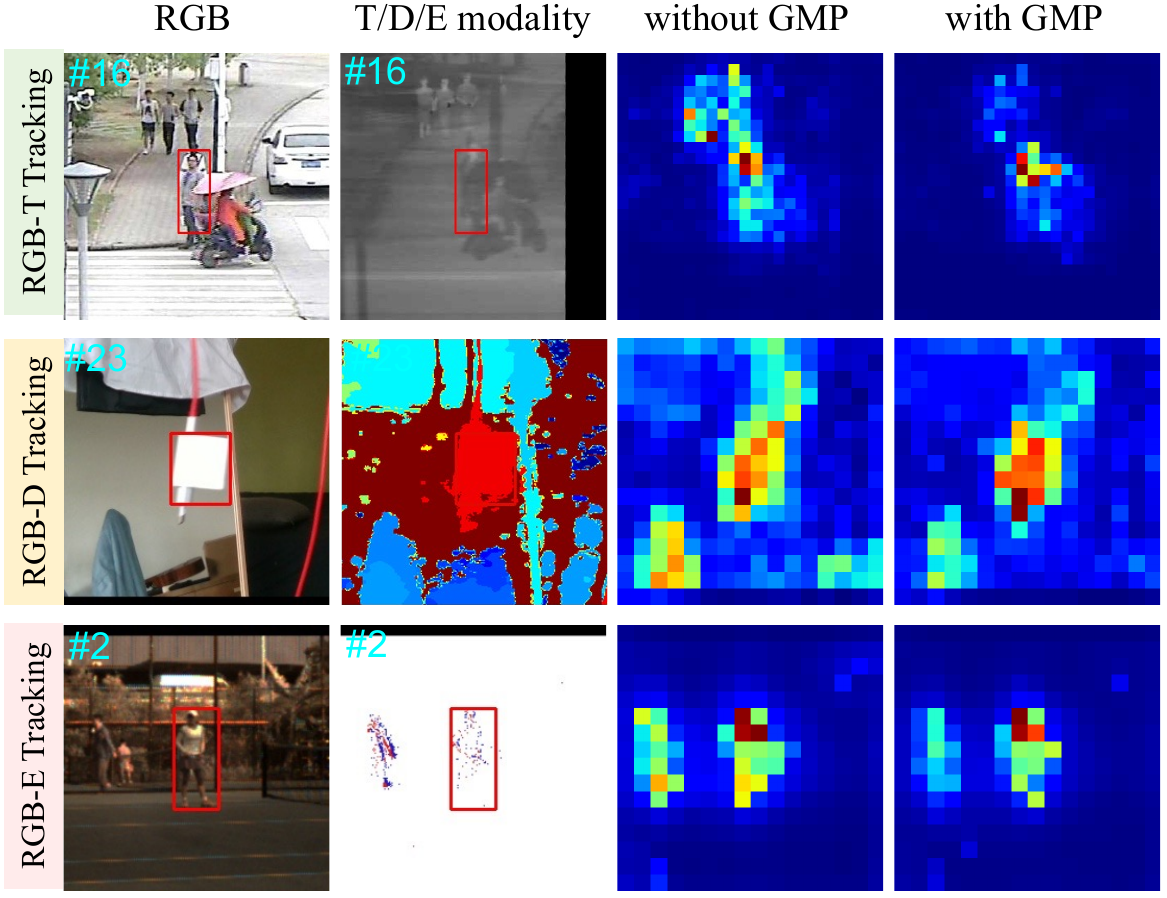}
       \caption{Feature visualization with and without the gated modal-scalable perceiver (GMP) on LasHeR, DepthTrack and VisEvent benchmarks.
       }
       \label{fig:gmp_feature}
    \end{figure}

\begin{table}[t]
    \centering
    \caption{Ablation study of search video-clip length on LasHeR \cite{lasher}, DepthTrack \cite{depthtrack}, and VisEvent \cite{visevent}. Results are reported in percentage (\%).}
    \label{tab:seq_length}
    \resizebox{\linewidth}{!}{
    \begin{tabular}{c|c|ccc|cc|cc}
    \toprule
    \multicolumn{1}{c|}{\multirow{2}{*}{\#}} & \multicolumn{1}{c|}{\multirow{2}{*}{Length}} & \multicolumn{3}{c|}{DepthTrack} & \multicolumn{2}{c|}{LasHeR} & \multicolumn{2}{c}{VisEvent} \\
     \cline{3-9}
     & & F-score($\uparrow$) & Re($\uparrow$) & Pr($\uparrow$) & SR($\uparrow$) & PR($\uparrow$) & SR($\uparrow$) & PR($\uparrow$) \\
     \midrule
     1 & \textit{2} & 69.1 & 70.6 & 67.7 & 60.1 & 75.4 & 60.8 & 78.6 \\
     2 & \textit{3} & \textbf{69.3} & \textbf{70.7} & \textbf{67.8} & \textbf{60.8} & \textbf{76.8} & \textbf{62.4} & \textbf{81.3} \\
     3 & \textit{4} & 68.6 & 70.1 & 67.3 & 59.4 & 74.4 & 61.5 & 79.1 \\
    \bottomrule
    \end{tabular} }
\end{table}

\textbf{The importance of multi-modal cues.} 
To validate the effectiveness of fusing RGB frames with frames from other modalities in visual tracking, we report the tracking results of RGB-only and bimodal data on the {\modaltracker}.
As shown in Tab.\ref{tab:multimodal_study} (\#1 and \#5), when only RGB frames are used, our tracker exhibits a significant performance drop across three downstream benchmarks. For instance, in the LasHeR dataset, the SR and PR metrics decrease by 6.3\% and 7.2\%, respectively, demonstrating that the injection of multi-modal cues (or multi-modal fusion) is notably effective and crucial for multi-modal tracking.

\textbf{The importance of shared modality tokenizers.}
We compared the impact of shared and non-shared tokenizers on multi-modal tracking performance. In the experiment, we encoded multi-modal data using both shared and non-shared tokenizers, and fed the encoded data into the tracking model for training and inference. A shared tokenizer refers to using a unified tokenizer to encode data from all modalities, while a non-shared tokenizer involves using different tokenizers for each modality. As shown in the Tab.\ref{tab:multimodal_study} (\#2 and \#5), we found that the tracking performance of the shared tokenizer was superior. This suggests that the shared tokenizer can more effectively capture the correlation between multi-modal data, thereby improving the overall performance of our tracker.

\textbf{Full fine-tuning $v.s.$ Adapter/Prompt fine-tuning.}
As shown in the Fig.\ref{fig:adapter}, we explore an experiment where different training strategies are used to train our model, such as adapter fine-tuning and full fine-tuning.
The experimental results are recorded in Tab.\ref{tab:multimodal_study} (\#3 and \#5).
We can be observe that both the \textit{adapter} and \textit{full} fine-tuning strategies achieve good performance improvements, with the \textit{full} fine-tuning showing slightly better results. 
In theory, adapter fine-tuning with fewer learning parameters can save more GPU resources. However, in practice, due to the gradients of other model parameters being retained during training, its training resources are not significantly reduced, which is comparable to the full fine-tuning scheme.
Therefore, we choose full fine-tuning with more learning parameters as our training strategy.

\textbf{Multi-task one-shot training $v.s.$ Single-task independent training.}
To estimate the benefits of multi-task unified (one-shot) training on our final model, we independently train expert models for three sub-tracking tasks, as shown in the Tab.\ref{tab:multimodal_study}. The comparison results between \#4 and \#5 show that our one-shot training scheme brings significant performance improvement. For example, the model trained solely on the DepthTrack dataset achieves an F-score of 67.8\%, whereas the model jointly trained on DepthTrack, LasHeR, and VisEvent achieves a higher F-score of 69.3\%, representing a 1.5\% improvement.
We attribute this improvement to the increased availability and diversity of training data for each modality tracking task, as well as the effectiveness of the designed gated perceivers in aggregating multi-modal features. Together, these factors enhance the robustness and generalization capability of our unified multi-modal tracking model across various tracking scenarios.

\begin{table}[t]
\centering
\caption{Comparison of model parameters, FLOPs, and inference speed.}
\resizebox{\linewidth}{!}{
\begin{tabular}{l|cccccc}
\toprule
Method & Type & Resolution & Params & FLOPs & Speed  & Device \\
\midrule
SeqTrack & ViT-B & $384\times384$ & 89M & 148G & 11$fps$ & 2080Ti \\
{\tracker} & ViT-B & $384\times384$ & 92M & 73G & 32$fps$ & 2080Ti \\
\bottomrule
\end{tabular} }
\label{tab:param}
\end{table}

\textbf{Modality weight ratios.}
we conducted a comparative study, shown in Tab.15, to evaluate the impact of different modality weights on model performance. As illustrated in the results, the tracker variants perform consistently well under different weighting schemes. For example, with a weight configuration of depth: infrared: event = 2:1:1, our model achieves F-score and SR values of 69.1\%, 60.4\%, and 61.7\% on the DepthTrack, LasHeR, and VisEvent datasets, respectively. These findings suggest that our method is not highly sensitive to the specific weighting of modalities. Therefore, the equal weighting scheme (depth: infrared: event = 1:1:1) can effectively balance the contributions from each modality and serves as a robust default configuration for our model.

\subsection{Qualitative Analysis}

\textbf{Speed, FLOPs and Params Analysis.}
We conduct comparative experiments in model parameters, FLOPs and inference speed, as shown in Tab.\ref{tab:param}.
With the same test machine (i.e., 2080Ti), our {\tracker} obtains faster inference speed compared to the latest tracker SeqTrack. And our tracker runs at 32 $fps$.

\textbf{Visualization.}
For RGB tracking task, to intuitively show the effectiveness of our method, especially in complex scenarios including similar distractors, we visualize the tracking results of our {\tracker} and three advanced trackers on LaSOT. As shown in Fig.\ref{fig:visual}, due to its ability to densely propagate trajectory information of the target, our tracker far outperforms the latest tracker SeqTrack on these sequences.

For multi-modal tracking tasks, we also visualize the multi-modal tracking results of our {\modaltracker} and other SOTA trackers on LasHeR, DepthTrack, and VisEvent datasets, respectively, as shown in Fig.\ref{fig:modal_visual}.
Benefiting from the universal awareness capability of the gated perceivers for arbitrary modalities, our {\modaltracker} is able to accurately localize target in complex sequences compared to other multi-modal trackers.
At the same time, we compare the feature representations with and without the gated modal-scalable perceiver (GMP). As shown in Fig.\ref{fig:gmp_feature}, without the GMP module, the model lacks the ability to capture inter-modal correlations, leading to learned representations that often attend to distractors similar to the target. In contrast, when equipped with the GMP module incorporating an attention-based gating mechanism, our tracker effectively suppresses such interference in complex multi-modal tracking scenarios, enabling the model to focus more accurately on the target object.

Furthermore, we visualize the attention map of temporal token attention operation, as shown in Fig.\ref{fig:attnmap}.
We can observe that the temporal token continuously propagate and attend to motion trajectory information of object, which aids our tracker in accurately localizing target instance.

\section{Conclusion}
In this work, we have explored an interesting video-level framework for visual object tracking, called {\tracker}. We reformulate visual tracking as a token propagation task that densely associates the contextual relationships of across video frames in an auto-regressive manner.
Furthermore, to extend from single-modal to multi-modal perception, we have proposed {\modaltracker}, a universal video-level modality-awareness visual tracking framework that effectively aggregates multi-modal temporal information of target instance through the design of gated attention mechanism.
Specifically, we have designed a video sequence sampling strategy and two temporal token propagation attention mechanisms, enabling the proposed framework to simplify video-level spatio-temporal modeling and avoid intricate online update strategies.
Moreover, we have proposed two gated modal-scalable perceivers to aggregate target spatio-temporal information from various modalities.
Finally, our model can inference different multi-modal tracking tasks simultaneously using the same set of model parameters through a one-shot training scheme.
Extensive experiments show that our {\modaltracker} achieves promising results on seven visible tracking and five multi-modal tracking benchmarks.
We expect that {\tracker} and {\modaltracker} will serve as powerful baselines for universal video-level modal-awareness tracking, inspiring further research in visible tracking and multi-modal tracking.

\textbf{Limitation.}
This work models the entire video as a sequence and decodes the localization of an object frame by frame in an auto-regressive manner. Despite achieving remarkable results, our video-level modeling method is a global approximation due to constraints in GPU resources, and we are still unable to construct the framework in a cost-effective manner. A promising solution would involve improving the computational complexity and lightweight modeling of the transformer.

\ifCLASSOPTIONcompsoc
  \section*{Acknowledgments}
\else
  \section*{Acknowledgment}
\fi

This work is supported by the National Natural Science Foundation of China (No.U23A20383, 62472109 and 62466051), the Project of Guangxi Science and Technology (No.2024GXNSFGA010001, 2025GXNSFAA069676, and 2025GXNSFAA069417), the Guangxi “Young Bagui Scholar” Teams for Innovation and Research Project, and the Research Project of Guangxi Normal University (No.2024DF001).

\ifCLASSOPTIONcaptionsoff
  \newpage
\fi



\bibliographystyle{IEEEtran}
\bibliography{IEEEabrv,reference}
%



\end{document}